\documentclass[sn-mathphys,Numbered]{sn-jnl}
\usepackage{graphicx}%
\usepackage{multirow}%
\usepackage{amsmath,amssymb,amsfonts}%
\usepackage{amsthm}%
\usepackage{mathrsfs}%
\usepackage{mathtools}
\usepackage[title]{appendix}%
\usepackage{xcolor}%
\usepackage{textcomp}%
\usepackage{manyfoot}%
\usepackage{booktabs}%
\usepackage{algorithm}%
\usepackage{algorithmicx}%
\usepackage{algpseudocode}%
\usepackage{listings}%
\usepackage{todonotes}%
\usepackage{bbm}%
\usepackage{wasysym}%
\usepackage{subcaption}
\usepackage{cleveref}

\usepackage{clipboard} 

\usepackage{multirow}

\theoremstyle{definition}
\newtheorem*{example*}{Example}

\theoremstyle{definition}
\newtheorem{example}{Example}

\theoremstyle{definition}

\theoremstyle{definition}
\newtheorem*{theorem*}{Theorem}

\theoremstyle{definition}
\newtheorem*{claim*}{Claim}

\theoremstyle{definition}
\newtheorem{theorem}{Theorem}

\theoremstyle{definition}

\theoremstyle{definition}

\theoremstyle{definition}
\newtheorem{corollary}{Corollary}

\theoremstyle{definition}
\newtheorem*{corollary*}{Corollary}

\theoremstyle{definition}
\newtheorem{definition}{Definition}

\theoremstyle{definition}

\theoremstyle{definition}
\newtheorem{lemma}{Lemma}

\theoremstyle{definition}
\newtheorem*{lemma*}{Lemma}

\newenvironment{delayedproof}[1]
 {\begin{proof}[\raisedtarget{#1}Proof of \Cref{#1}]}
 {\end{proof}}
\newcommand{\raisedtarget}[1]{%
  \raisebox{\fontcharht\font`P}[0pt][0pt]{\hypertarget{#1}{}}%
}
\newcommand{\proofref}[1]{\hyperlink{#1}{Appendix}}

\newcommand{\indep}{\rotatebox[origin=c]{90}{$\models$}}
\newcommand{\RR}{\mathbb{R}}

\newcommand{\NN}{\mathbb{N}}

\newcommand{\chara}{\mathbbm 1}
\newcommand*\diff{\mathop{}\!\mathrm{d}}

\DeclareMathOperator{\diag}{diag}
\DeclareMathOperator{\ips}{IPS}
\DeclareMathOperator{\roc}{ROC}

\DeclareMathOperator{\symratedependent}{\mathcal{R}}
\newcommand{\realratedependent}[2]{\widehat{\symratedependent} ({#2}, {#1})}

\newcommand{\REval}[2]
{
	\expandafter\ifx\expandafter\relax\detokenize{#2}\relax
	{
		\expandafter\ifx\expandafter\relax\detokenize{#1}\relax
		{
			{\mathcal{R}}
		}
		\else{
			{\REval{}{}}({#1})
		}\fi	
	}
	\else
	{
		\REval{}{}({#1}, {#2})
	}\fi	
}

\newcommand{\EEval}[2]
{
	\expandafter\ifx\expandafter\relax\detokenize{#2}\relax
	{
		\expandafter\ifx\expandafter\relax\detokenize{#1}\relax
		{
			{\mathcal{E}}
		}
		\else{
			{\EEval{}{}}({#1})
		}\fi	
	}
	\else
	{
		\EEval{}{}({#1}, {#2})
	}\fi	
}

\newcommand{\FEval}[2]
{
	\expandafter\ifx\expandafter\relax\detokenize{#2}\relax
	{
		\expandafter\ifx\expandafter\relax\detokenize{#1}\relax
		{
			{\mathcal{F}}
		}
		\else{
			{\FEval{}{}}({#1})
		}\fi	
	}
	\else
	{
		\FEval{}{}({#1}, {#2})
	}\fi	
}

\DeclareMathOperator{\femmina}{{\text{\female}}}
\DeclareMathOperator{\maschio}{{\text{\male}}}

\newcommand{\agroup}[1]
{
	\ifnum\numexpr#1\relax=0%
		{\femmina}%
	\else%
		{\maschio}%
	\fi%
}

\newcommand{\varrate}[1]
{
	\ifnum\numexpr#1\relax=0%
		{\eta}%
	\else%
		{\rho}%
	\fi%
}

\newcommand{\varfix}[1]
{
	\ifnum\numexpr#1\relax=0%
		{\overline\eta}%
	\else%
		{\overline\rho}%
	\fi%
}

\newcommand{\jointrate}[3]
{
\expandafter\ifx\expandafter\relax\detokenize{#3}\relax
	{
		\prob({#2}={#1}, \labely ={#1})
	}
	\else{
		\prob({#2}={#1}, \labely ={#1}\conditional {#3})
	}\fi
}

\newcommand{\paretoless}{\lesssim}
\newcommand{\paretomore}{\gtrsim}
\newcommand{\paretoeq}{\sim}
\newcommand{\paretostrictless}{<}
\newcommand{\paretostrictmore}{>}

\newcommand{\symmeasure}{\mu}
\newcommand{\featureset}{X}
\newcommand{\individual}{x}

\newcommand{\prob}{P}
\newcommand{\slicing}{S}
\newcommand{\labely}{\mathbf{y}}
\newcommand{\labeloutcome}{{y}}
\newcommand{\modeloutcome}{\widehat{\labeloutcome}}
\newcommand{\model}{{\widehat{\labely}}}
\newcommand{\labelset}{Y}
\newcommand{\binaryset}{\{0,1\}}
\newcommand{\optimalmodel}{\overline{\labely}}
\newcommand{\vectorize}[1]{\boldsymbol{#1}}

\newcommand{\conditional}{\mid}
\newcommand{\restricted}[1]{\big\vert_{#1}}

\DeclareMathOperator{\EE}{\mathbb{E}}
\DeclareMathOperator*{\EEsub}{\EE}

\usepackage[autostyle=false, style=english]{csquotes}
\MakeOuterQuote{"}

\newcommand{\FairnessProblem}[1]
	{
	\arg\max_{\model\colon \featureset\to \binaryset} {#1}
	\EEval{\model}{}-
	 \left\vert
		\FEval{\model\restricted{\agroup{0}}}{} - \FEval{\model\restricted{\agroup{1}}}{}
	\right\vert
	}
\usepackage{anyfontsize}
\begin{document}

%


\title{
	\centering
	Cherry on the Cake:\\ 
	Fairness is NOT an Optimization Problem
}

\author*[1]{\fnm{Marco} \sur{Favier}}\email{marco.favier@uantwerpen.be}

\author[1]{\fnm{Toon} \sur{Calders}}

\affil[1]{\orgname{University of Antwerp}, \orgaddress{\city{Antwerp}, \country{Belgium}}}

\abstract{
	In Fair AI literature, the practice of maliciously creating unfair models that nevertheless satisfy fairness constraints is known as \emph{cherry-picking}. A cherry-picking model is a model that makes mistakes on purpose, selecting bad individuals from a minority class instead of better candidates from the same minority. The model literally cherry-picks whom to select to superficially meet the fairness constraints while making minimal changes to the unfair model.
	This practice has been described as "\emph{blatantly unfair}" and has a negative impact on already marginalized communities, undermining the intended purpose of fairness measures specifically designed to protect these communities.
	A common assumption is that cherry-picking arises solely from malicious intent and that models designed only to optimize fairness metrics would avoid this behavior.
	We show that this is \emph{not} the case: models optimized to minimize fairness metrics while maximizing performance are often forced to cherry-pick to some degree. In other words, cherry-picking might be an inevitable outcome of the optimization process itself.
	To demonstrate this, we use tools from fair cake-cutting, a mathematical subfield that studies the problem of fairly dividing a resource, referred to as the "cake," among a number of participants.
	This concept is connected to supervised multi-label classification: any dataset can be thought of as a cake that needs to be distributed among different labels, and the model is the function that divides the cake.
	We adapt these classical results for machine learning and demonstrate how this connection can be prolifically used for fairness and classification in general.
}
\maketitle
\section{Introduction}\label{sec_introduction}

In the field of \emph{Fair ML} \cite{barocas2023fairness}, researchers focus on developing machine learning algorithms and AI models that minimize bias and discrimination, ensuring equitable outcomes for all individuals, regardless of their demographic characteristics.
To measure the difference in decision between communities, various mathematical metrics have been developed to evaluate the extent of unfairness in a classifier. For \emph{group fairness}, in particular, where a model is considered fair when different demographic groups are treated similarly, the fairness criterion is often connected to the performance of the classifier in question.

Fairness metrics are often used as additional objectives in optimization problems, but the approach has been criticized for its naivety \cite{Wachter2020WhyFC, Binns2018ItsRA, favier2023fair, weerts2022does}. Moreover, it remains unclear how the fairness term affects the learning process or the final model \cite{wick2019unlocking, haas2019price, Gerrymandering_pmlr-v80-kearns18a}, and whether these models can even be considered fair.
Despite these concerns, no mathematical proof has been provided to justify the general skepticism surrounding the optimization of fairness metrics, and the few existing formal results are limited to very specific fairness constraints \cite{corbett2017algorithmic, menon2018cost, baumann2022enforcing,hardt2016equality}.

This paper provides an important forward step on this issue, mathematically proving that optimizing group fairness metrics can frequently exhibit problematic behaviors, wherein the optimal solution inherently possesses unfair properties.
We show that when fairness is framed as an optimization problem, it is often optimal to deliberately introduce errors in a specific demographic group. 

This issue differs fundamentally from the necessity of making different decisions across demographic groups to satisfy fairness constraints, a situation often driven by pre-existing bias or discrimination in the data. Instead, the phenomenon we describe is distinct: the model intentionally makes mistakes within the same community, even when it is unnecessary to do so to satisfy the fairness constraint. The model favors candidates with low predicted performance over those with high predicted performance, even when both candidates belong to the same demographic group.

The careless, if not malicious, selection of individuals solely to meet group fairness constraints is often referred to in the literature as \emph{cherry-picking} \cite{cherrypick_fleisher2021s, cherrypick2_dwork2012fairness, cherrypick3_goethals2023precof}.
Cherry-picking is not only inherently unfair, as it marginalizes worthy individuals from minorities, but also risks perpetuating stereotypes and biases, as less-capable individuals will be selected by the model.

For example, a model optimized to avoid ageism in the workplace, given two young candidates of the same age, might paradoxically prefer to hire the worst one of the two on purpose. If a worker in the same company wrongly assumes that newer generations are lazy, this unfair selection would offer anecdotal evidence for their prejudgment, as the newly hired person will likely underperform. The model is essentially fueling confirmation bias, which reinforces the unfair stereotyping of young individuals.


Generally, cherry-picking is portrayed as a mischievous and absurd method, capable of unfairly meeting fairness constraints, that should be avoided.
Our results, however, show that cherry-picking can be a phenomenon that inevitably emerges from the optimization process itself. Unaware practitioners might thus obtain a cherry-picking model without even realizing it.

We prove this by using fair cake-cutting theory, a mathematical tool that studies how a resource can be distributed. In various social and economic contexts, fairly allocating a divisible resource among multiple participants is of paramount importance. When the resource is heterogeneous in nature, the players involved may even disagree on the value of a piece of it.

For instance, consider the problem of dividing land for agricultural purposes. The yield of different crops depends on various factors, including the chemical components in the soil, access to sunlight and water, and the overall geography of the land. For this reason, different plots of land may be more suitable for some crops than for others. Efficiently allocating the land is imperative to maximize the total yield. In this example, the different crops can be thought of as players demanding a certain amount of land, each disagreeing about the value of a particular plot.
How to divide and distribute a heterogeneous resource (the land) among different participants (the crops) is the kind of question that cake-cutting theory aims to solve.
%
%

In multi-label supervised learning, researchers must address a similar problem: learning a model capable of predicting a label for each data point in a dataset.
The model's prediction essentially partitions the dataset, where each partition corresponds to a predicted label. From a cake-cutting perspective, each label acts as a player: the more data points assigned to a specific label, the higher the prediction rate for that label becomes.

For example, in binary classification, assigning the positive label to data points previously labeled as negative results in an equal or increased true positive rate.
Furthermore, the dataset, or more precisely, the underlying distribution, behaves as a heterogeneous resource: the likelihood of each label varies from data point to data point. Hence, every label \emph{prefers} some data points more than others.

Under some reasonable assumptions, not only are the two problems related, but they are actually equivalent: every cake-cutting problem can be expressed as a multi-label classification problem, and vice versa.
Because of this connection, a number of results from cake-cutting theory can be restated using multi-label classification terminology. This translation sheds light on the space of possible decisions and feasible model performances, generalizing concepts such as $\roc$ curves and confusion matrices.
%
%
This connection was already established in the past, but to the best of our knowledge, it has not been used in the context of machine learning or fairness.

We use this connection to prove our claims that blindly optimizing fairness metrics will, in many cases, result in counterintuitive and unfair outcomes, as optimal models may be forced to cherry-pick. We strongly believe that these findings highlight why discussing fairness purely as a mathematical optimization problem is disingenuous and potentially dangerous.
%
\section{Understanding Cherry-Picking}\label{sec_understanding_cherry}
Consider the following illustrative scenario:

\medskip
\begin{example}\label{example:cherry_picking}
You have developed a new ML model $\model\colon \featureset\to \binaryset$ for college applications, able to predict if a given applicant $\individual\in \featureset$ would be a good student $(\labely =1)$ or drop out $(\labely =0)$. Despite your best efforts, the model seems to perform differently depending on the gender $A\in\{\agroup{0},\agroup{1}\}$ of the considered applicant.

In particular, the precision of the model for men $\prob( \labely = 1\mid \model = 1, A=\male)$ appears conveniently low. This has raised concerns about the fairness of the model, as the difference in precision suggests that the model is very strict when selecting which female candidates to admit to college, while being more lenient with male candidates. You know that, by law, unfair treatment of different individuals based on gender is not allowed, and that the precision needs to be similar for all groups.

Nonetheless, your model has been recently adopted by a conservative institution that is aware that the performance needs to change to abide by the law. You have been contacted to discuss how to proceed with the matter.

The university does not want to increase the precision for men: higher precision would mean denying education to multiple candidates, possibly excluding many potentially good students. You agree that denying education to anyone is not an option you want to pursue.

You then propose lowering the precision for women to admit more of them. But to your surprise, the university staff does not appreciate this idea: they claim they are quite overworked, and remind you that any new student means a little extra labor for them. It's pretty clear that the university doesn't really care about the fairness of the model or about gender equality. To the point that they propose a murky deal instead. You have two options: 
\begin{enumerate} 
	\item You do the ethical thing and lower the decision threshold for female applicants, so that women who were previously on the verge of being selected are now admitted. This group still likely contains many suitable candidates. Consequently, you would have to admit a larger number of them before achieving an equal precision rate as that of males.
	
	\item You accept the university's proposal: they suggest tweaking the model to admit only a few female students who would clearly drop out, purposefully making mistakes. The number of additional students admitted would be kept to a minimum, all of whom would drop out soon anyway. This approach would indeed be sufficient to balance the errors and provide the same precision for both groups.
	This option is highly unethical, but the university doesn't care. They only care to show that they are abiding by the law and that the precision is now `fair'. They completely wash their hands of the matter: according to them, they are still giving women an opportunity, it's not their fault if they fail.
\end{enumerate}
\end{example}
\medskip
In the field of fairness in ML, researchers deal with the problem of creating predictive models $\model\colon \featureset\to \labelset$, which have the additional requirement of "being fair" to different subgroups of the population, called \emph{sensitive groups}. In our example, the sensitive groups correspond to different genders. The elusive notion of fairness is often defined within the problem and evaluated using a fairness metric.

In the context of our example, the definition of fairness was that the model's performance needed to be the same for men and women. Fairness constraints that depend on how a classifier behaves on different subgroups of the population are often called \emph{group fairness constraints} to emphasize that the fairness concept they aim to enforce can be measured and understood only by looking at how a classifier performs on the entire population. In contrast, \emph{individual fairness} explores concepts of fairness applied to single individuals. Our work focuses on the former.

The two possible solutions proposed in the example are two different ways to satisfy the fairness constraint. The first solution is, in our view, the correct ethical solution, where different thresholds are used for different groups to guarantee a fair outcome. The second solution, on the other hand, is an unethical way to circumvent fairness constraints by intentionally making mistakes, a practice referred to as \emph{cherry-picking} in the fairness literature \cite{cherrypick2_dwork2012fairness, cherrypick_fleisher2021s, cherrypick3_goethals2023precof}. 
Formally, we define cherry-picking as follows: 
\medskip 
\begin{definition}[Cherry-Picking] We say that a probabilistic model $\model\colon \featureset\to \labelset$ trained to predict a binary label $\labelset\in\binaryset$ does not \emph{cherry-pick} if, for all individuals $\individual, \individual'$ in the same sensitive group $A$, the following condition holds: 
	\begin{equation*} 
		\text{If } \prob(\labely = 1 \mid \individual) < \prob(\labely = 1 \mid \individual'), \text{ then } \prob\left(\model(\individual') = 0 \mid \model(\individual) = 1 \right) = 0 
	\end{equation*} If this condition does not hold, we say that $\model$ \emph{cherry-picks}. 
\end{definition} \medskip
Let's unpack why a model that doesn't satisfy the previous definition can be considered cherry-picking and deemed unfair.
If a model $\model$ cherry-picks, it means that we are able to find individuals $\individual, \individual'$ from the same sensitive group $A$ such that, even if $\individual'$ better deserves to receive the positive label than $\individual$, the model $\model$ will still assign the positive label to $\individual$ and not to $\individual'$.

This means that the model is selecting individuals not based on their merits, but solely to satisfy certain fairness constraints. It's important to note that this is different from the case where $\individual$ and $\individual'$ belong to different sensitive groups. In that case, the difference in outcomes can be more easily justified. For instance, the distribution of the two groups may be skewed due to societal biases. Therefore, it's acceptable if the prediction for the individual in the privileged group is negative, while the one for the person in the unprivileged group is positive, in order to compensate for the bias.
Alternatively, affirmative action policies may require the use of different criteria when assigning the positive label to the two groups

This example, albeit unrealistic, illustrates the conceptual questions that fair ML practitioners have to face: balancing performance and fairness, and choosing the fairest solution among different options.

In this paper, we specifically focus on unethical predictive models similar to the second one in the example. Specifically, we investigate instances where cherry-picking solutions may arise naturally from the definition of the fairness problem, despite being unethical. In other words, we explore situations where optimal solutions to fairness problems are not fair at all.

\section{Related Works and Contributions}\label{sec_related_works}

\subsection{On Fairness in Machine Learning}
In Section~\ref{sec_understanding_cherry}, Example~\ref{example:cherry_picking} illustrates that determining an optimal model under fairness constraints can be far from straightforward. To address such questions, Corbett-Davis et al. \cite{corbett2017algorithmic} have considered how to find optimal models when utilizing the following fairness constraints: 
\begin{itemize} 
	\item (Conditional) Statistical Parity: $\prob( \model=1 \conditional A)=\prob( \model=1)$\\ Which means that individuals should have the same access to the positive label, independent of the sensitive group $A$ (for the "conditional" case, subgroups that satisfy the same requirements are used). 
	\item Equal Opportunity: $\prob( \model=1 \conditional \labely = 1, A)=\prob( \model=1 \conditional \labely = 1)$\\ 
	Which means that the model's errors on individuals who deserve the positive label should be independent of the sensitive group to which they belong. 
	\item Equal Risk: $\prob( \model=0 \conditional \labely = 0, A=\agroup{1})=\prob( \model=0 \conditional \labely = 0)$\\ 
	Which is the same as Equal Opportunity, but for the negative label. 
\end{itemize} 
They proved that, if models are evaluated according to "Immediate Utility," an optimal fair model can be attained by employing distinct thresholds on the probability distribution for each sensitive group. 
Immediate Utility is defined as follows: 
\begin{equation*}\label{eq:im utility} 
	{U}_t(\model):=\prob( \model= 1,\labely = 1) - t\cdot \prob( \model= 1) 
\end{equation*} 
where $t$ is a constant within the range $[0, 1]$. They limit their results to the case where fairness is a constraint to satisfy, rather than the more general case of a measure to minimize.

Menon et al. \cite{menon2018cost} further improved on these results by showing that when the aforementioned fairness requirements are implemented as part of the maximization problem by employing a fairness metric, the optimal classifier can still be found with different thresholds on the probability distribution. However, their work, like that of Corbett-Davis et al., is limited to Immediate Utility.

Our contribution generalizes these findings to encompass any evaluation metric. More specifically, Theorem \ref{teo_when_works} introduces a general tool to check if an optimal solution can be found via threshold optimization on each sensitive group. We show how this tool can be used on the measures considered by Corbett-Davis et al. in Theorem \ref{teo_dpd_eo_ect}.

The importance of these results is twofold: first, they demonstrate that some of the earliest methods for achieving fairness are not only still relevant but also theoretically optimal \cite{kamiran2012data, hardt2016equality}. This is significant since it shows that the best way to achieve fairness is to train a well-calibrated classifier, able to correctly assign probability distributions to different individuals, instead of using a more complex, and often less interpretable, model. 

Secondly, these results can be used to measure the so-called "fairness-accuracy trade-off." The fairness-accuracy trade-off suggests that an unconstrained model will outperform one that satisfies a fairness constraint \cite{corbett2017algorithmic, menon2018cost, pleiss2017fairness}. The intuition behind why this phenomenon occurs can be easily understood: the set of constrained models is a subset of all possible models, and hence there is a high probability that the best model exists outside the constrained set. However, this point of view has also been challenged \cite{wick2019unlocking, favier2023fair}: researchers have shown that fair models are better suited to predict unbiased data after introducing errors in the dataset through a biasing procedure. Under this framework, the trade-off becomes an illusory consequence caused by the bias in the data.

\subsection{On Cherry-Picking}
One important overlooked consequence of what Corbett-Davis et al. and Menon et al. have proven is that, for some specific fairness problems, it is not necessary to cherry-pick to find optimal solutions.

Cherry-picking is often discussed to underline that group fairness constraints are not enough to ensure the fairness of a model. The overloading of some terms often creates confusion in that regard: in machine learning terminology, we often say that a model is fair if it satisfies the chosen fairness constraint, but this does not mean that the model is fair in the broader ethical sense.

Cherry-picking is a clear example that shows why it is not sufficient to exhibit a low (un)fairness measure to claim having a fair model, as it has been described as "blatantly unfair" while still being able to satisfy group fairness constraints \cite{cherrypick2_dwork2012fairness, cherrypick_fleisher2021s}.

What is often overlooked is that cherry-picking can emerge as a consequence of the choice of the fairness constraint and the evaluation used. In particular, optimal "fair" models may be forced to cherry-pick.

Hardt et al. \cite{hardt2016equality} were the first to show that there are cases where cherry-picking is necessary to satisfy group fairness constraints. They showed that, for the case of Equal Odds, unless the $\roc$ curves of the two sensitive groups intersect, it is impossible to find a fair model that does not cherry-pick. It is also interesting to note that in their work, they had to prove properties about the space of possible decisions, which would be automatically granted by cake-cutting theory, showing how much the two fields are intertwined and why we believe it is important to discuss them.

More recently, Baumann et al. \cite{baumann2022enforcing} have shown that Predictive Parity, which requires a model to have the same precision on each sensitive group, can also lead to cherry-picking when paired with Immediate Utility in the optimization objective.

We greatly expand on the results from \cite{hardt2016equality, baumann2022enforcing} in Theorem \ref{teo_bad_fairness}, showing that for a wide range of fairness constraints, cherry-picking may be an unavoidable consequence of the optimization process.

Theorem \ref{teo_bad_fairness} can also be interpreted as a new impossibility result. Fairness literature has shown that when different ethical requirements are mathematically defined into fairness constraints, it's often impossible to satisfy them all \cite{friedler2021possibility,chouldechova2017fair, barocas2023fairness, bell2023possibility, Kleinberg2016InherentTI}. If we consider non-cherry-picking as an ethical requirement, then Theorem \ref{teo_bad_fairness} shows that it can be impossible to satisfy it while also optimizing for a group fairness constraint.

Impossibility results make it clear that choosing the right fairness measure is crucial. Understanding fairness as a philosophical, legal, and social science problem \cite{binns2018fairness, Binns2018ItsRA, veale2018fairness, carey2023statistical, carey2022causal, Wachter2020WhyFC} is a necessity for any practitioner who is interested in implementing a fairness constraint, as the choice has practical, legal, and ethical consequences.

\subsection{On Cake-Cutting}
Cake-cutting encompasses a wide range of different problems. Some approaches are more algorithmic in nature, where the goal is to find a procedure to divide the cake fairly \cite{brams1995envy, aziz2016discrete}. Others are more theoretical, focusing on the properties of the set of possible solutions \cite{weller1985fair}. Our work is more closely related to the latter, explaining how theoretical results from cake-cutting find equivalent results in classification, especially in fairness. The connection between cake-cutting and statistical decision-making was first explored by Dvoretzky et al. in \cite{dvoretzky1951elimination}.

Our only formal contribution to cake-cutting is limited to Theorem~\ref{teo_extra_dim}, which is a slightly different version of a result from \cite{dvoretzky1951relations}. All the other results for cake-cutting are not new, yet we believe we are the first to use them in the context of machine learning and its terminology. 
Theorem \ref{teo_IPS_open}, Lemma \ref{lemma_segment}, and Corollary \ref{cor_binary_weller} are all established results that do not provide any new information about cake-cutting, but have never been stated to describe ML-specific concepts like aleatoric uncertainty, deterministic classifiers, or ROC curves. Theorem \ref{teo_atomic_cherry_picking} provides a characterization of cherry-picking models through cake-cutting that is completely new.

These results are an important tool for understanding the properties of classification problems. In particular, an in-depth explanation is provided in Section~\ref{sec_connection}, showing the connection between cake-cutting and machine learning multi-label classification.
\section{Notation \& Definitions}\label{sec_notation}
\subsection{Cake-Cutting Theory}
Cake-Cutting Theory formalizes problems where a divisible resource needs to be distributed among different players in an optimal way. Each player exhibits preferences on what portion of the resource they prefer to receive. It has applications in Game Theory and Statistics.
\medskip
\begin{definition}[Cake-cutting]
	A \emph{cake-cutting instance} is a triple $(\featureset ,\Sigma, \vectorize{\symmeasure})$ where
	\begin{itemize}
		\item $\featureset $ is a set called the "\emph{cake}," and each player wants a metaphorical slice of it.
		\item $\Sigma$ is a $\sigma$-algebra on $\featureset $, representing all the possible slices of the cake.
		\item $\vectorize{\symmeasure}:=(\symmeasure_1,\dots,\symmeasure_n)$ is a vector of $n$ atomless\footnote{
			A measure $\symmeasure$ is said to be \emph{atomless} if for each positive measurable set $A$ there exists $B\subset A$ such that $0<\symmeasure(B)<\symmeasure(A)$. Otherwise, $\symmeasure$ is said to have \emph{atoms}.
		} 
		finite measures on the measurable space $(\featureset ,\Sigma)$. Each measure represents a player who wants a piece of the cake, $\symmeasure_i$ is how much player $i$ likes a particular slice.
	\end{itemize}
\end{definition}
\medskip
\begin{definition}[Slicing]
	Given a cake-cutting instance $(\featureset ,\Sigma, \vectorize{\symmeasure})$, we call an ordered partition $\vectorize{\slicing}=(\slicing_1,\dots,\slicing_n)$ of $\featureset $ into $n$ measurable subsets a \emph{slicing}. 
	Each player $\symmeasure_i$ receives slice $\slicing_i$ and likes it $\symmeasure_i(\slicing_i)$. 
\end{definition}
\medskip
Obviously, players can also evaluate pieces received by other players.
This concept plays an important role in cake-cutting theory, so we need to keep track of the opinions of all the players for all the possible pieces.
\medskip
\begin{definition}[$\vectorize{\symmeasure}(\vectorize{\slicing})$ and $\ips$]
	Given a cake-cutting instance $(\featureset ,\Sigma, \vectorize{\symmeasure})$ and a slicing $\vectorize{\slicing}$, we indicate with $\vectorize{\symmeasure}(\vectorize{\slicing})$ the matrix defined as
	\begin{equation*}
		[\vectorize{\symmeasure}(\vectorize{\slicing})]_{i,j}:=\symmeasure_i(\slicing_j)\quad \forall i,j\in[n]
	\end{equation*}
	The \emph{Individual Pieces Set} $(\ips)$ of $\featureset $ is the set of possible diagonals of $\vectorize{\symmeasure}(\vectorize{\slicing})$, that is:
	\begin{equation*}
		\ips(\featureset ) :=\{\diag\vectorize{\symmeasure}(\vectorize{\slicing})\colon \vectorize{\slicing}\text{ is a slicing of }\featureset \}
	\end{equation*}
\end{definition}
So the matrix $\vectorize{\symmeasure}(\vectorize{\slicing})$ considers how players evaluate all the pieces in a given slicing, while the $\ips$ tracks how much each player likes their own piece, but does so for all possible slicings.
Since we want to split and distribute our cake so that every participant appreciates their slice, formally, it means that we would like to maximize $\symmeasure_i(\slicing_i)$. 
This naturally gives rise to the following definition:
\medskip
\begin{definition}[Pareto pre-order]
	Let $(\featureset ,\Sigma, \vectorize{\symmeasure})$ be a cake-cutting instance.
	Given two slicings $\vectorize{\slicing},\vectorize{\slicing}'$, we say that $\vectorize{\slicing}'$ is \emph{Pareto non-inferior} to $\vectorize{\slicing}$ (and similarly $\vectorize{\slicing}$ is \emph{non-superior} to $\vectorize{\slicing}')$, and write $\vectorize{\slicing}\paretoless\vectorize{\slicing}'$,
	if for all $i\in [n]$ it holds $\symmeasure_i(\slicing_i)\leq \symmeasure_i(\slicing'_i)$, or equivalently
	\begin{equation*}
		\diag\vectorize{\symmeasure}(\vectorize{\slicing})\leq \diag\vectorize{\symmeasure}(\vectorize{\slicing}')
	\end{equation*}
	We say that $\vectorize{\slicing}$ and $\vectorize{\slicing}'$ are \emph{equivalent}, and write $\vectorize{\slicing}\paretoeq\vectorize{\slicing}'$, if $\vectorize{\slicing}\paretoless\vectorize{\slicing}'$ and $\vectorize{\slicing}'\paretoless\vectorize{\slicing}$.\\
	A slicing $\vectorize{\slicing}$ is said to be \emph{Pareto-Optimal} if for any slicing $\vectorize{\slicing}'$ such that $\vectorize{\slicing}\paretoless \vectorize{\slicing}'$, it also holds $\vectorize{\slicing}'\paretoeq \vectorize{\slicing}$.	
\end{definition}
\subsection{Multi-Label Classification}
In supervised multi-label classification, we have a population of individuals $\featureset$ and a set of $n$ possible labels. Each individual is associated with a probability distribution over the labels, and our task is to find a probabilistic classifier that can predict the labels of new individuals.
\medskip
\begin{definition}
	We define a $(n$-dimensional) \emph{classification problem} on $\featureset $ as a probability space $\featureset \times \labelset$, where
	\begin{itemize}
		\item $\featureset $ is a set, called the \emph{feature space}, with probability measure $\symmeasure_\featureset$ and $\sigma$-algebra $\Sigma_\featureset$.
		\item $\labelset=\{1,\dots,n\} = [n]$ is the set of possible labels, which are sampled by a random variable $\labely $ having conditional probability density function
		\begin{equation*}
		\prob_{\featureset}(\labely \conditional -)\colon \featureset\to \Delta^n,\quad
		\individual\mapsto \prob_{\featureset}(\labely \conditional \individual)
		=
		(
			\prob_{\featureset}(\labely = 1\conditional \individual),\dots, \prob_{\featureset}(\labely = n\conditional \individual)
		)
		\end{equation*} where $\Delta^n:=\{\vectorize{t}\in [0,1]^n\colon t_1+\dots+t_n=1\}$ is the $n$-dimensional simplex. 
		We will write $\prob$ instead of $\prob_{\featureset}$ when it is clear from the context.
	\end{itemize}
	Our objective is to find a measurable probabilistic classifier (or model) {$\model\colon \featureset \to \labelset$} that we can use to make predictions.
	For a data point $\individual\in \featureset$, the classifier $\model(\individual)$ assigns a label to $\individual$ by sampling it according to a distribution $\prob(\model(\individual)= j )\in\Delta^n$. 
	We can understand how well a classifier performs by looking at the \emph{confusion matrix} $M(\model)$, which is defined as follows:
	\begin{equation*}
		M( \model)
		:=
		\left[\int_\featureset \prob(\model(\individual)= j)\prob(\labely = i\conditional \individual) \symmeasure_\featureset(\diff \individual)\right]_{i,j\in [n]}
		= 
		\left[\prob(\model= j, \labely = i)\right]_{i,j\in [n]}
	\end{equation*}

	Our objective is to maximize the \emph{prediction probabilities}, which correspond to the diagonal entries of the confusion matrix, by minimizing the \emph{error probabilities}, which are the entries that are not on the diagonal. A model $\model$ is said to be \emph{deterministic} if for all $\individual\in \featureset$, there exists an $i\in [n]$ such that $\prob(\model(\individual)=i)=1$. 
\end{definition} 
\section{Connecting Classification and Cake-Cutting}\label{sec_connection}
The connection between cake-cutting and supervised learning is intuitive: if we limit ourselves to deterministic models, then for any model $\model$, we can simply define a slicing $\vectorize{\slicing}$ so that
\begin{equation*}
	\slicing_j = \{\individual\in \featureset\colon \prob(\model(\individual)=j)=1 \}
\end{equation*}
and then define probability measures $\symmeasure_i$ by considering the entries of the confusion matrix, that is:
\begin{equation*}
	\symmeasure_i(\slicing_j)=\prob(\model= j, \labely = i) = {\int_\featureset \prob(\model(\individual)=j)\prob(\labely = i\conditional \individual) \symmeasure_\featureset(\diff \individual) }
\end{equation*}
Yet, it is unclear how we can define a proper slicing when the model used is not deterministic. 
%
When the measure $\symmeasure_\featureset$ is itself atomless, it is possible to prove that it doesn't really matter whether a model is deterministic or not: any confusion matrix obtained by a probabilistic decision can also be obtained by a deterministic one, as proven in \cite{dvoretzky1951relations}. 
\medskip
\begin{theorem}[Dvoretsky, Wald, and Wolfovitz's Theorem \cite{dvoretzky1951relations}]
	Consider a classification problem on $\featureset $ with atomless measure $\symmeasure_\featureset$. There exists a vector of measures $\vectorize{\symmeasure} = (\symmeasure_1,\dots, \symmeasure_n)$ such that the cake-cutting instance $(\featureset , \Sigma_\featureset, \vectorize{\symmeasure})$ satisfies:
	\begin{equation*}
		M = M( \model)\text{ for a model } \model \iff M=\vectorize{\symmeasure}(\vectorize{\slicing}) \text{ for some slicing } \vectorize{\slicing}=(\slicing_1,\dots, \slicing_n) 
	\end{equation*}
	for any matrix $M$.
\end{theorem}
\medskip
But for classification problems, we don't require $\symmeasure_\featureset$ to be atomless. This difference is not merely cosmetic, but it has important ramifications about cherry-picking models and deterministic solutions which are discussed in Section~\ref{subsec:the one cake}, after Lemma~\ref{lemma_segment}, and after Theorem~\ref{teo_atomic_cherry_picking}. 
To still preserve the connection between classification and cake-cutting, we prove our own version of the aforementioned theorem. What we prove is that any probabilistic model on $\featureset $, even when $\symmeasure_\featureset$ is not atomless, can be seen as a deterministic model on $\featureset \times [0,1)$, and hence a cake-cutting problem on $\featureset \times [0,1)$. Formally:
\medskip
\begin{theorem}\label{teo_extra_dim}\Copy{teo_extra_dim}{
	Consider a classification problem on $\featureset $. There exists a cake-cutting instance on $\featureset \times [0,1)$ such that for any matrix $M$, it holds:
	\begin{equation*}
		M = M( \model)\text{ for a model } \model \iff M=\vectorize{\symmeasure}(\vectorize{\slicing}) \text{ for some slicing } \vectorize{\slicing}=(\slicing_1,\dots, \slicing_n) 
	\end{equation*}
	}
	\begin{proof}
		In the \proofref{teo_extra_dim}.
	\end{proof}
\end{theorem}
The theorem directly implies that for each classification problem, there exists an equivalent cake-cutting instance. But the converse is also true: every cake-cutting problem can be stated as a classification problem, up to a reparametrization. This is a classical result \cite{density_exists}, which involves the use of the Radon-Nikodym theorem. Consider a cake-cutting instance $(\featureset ,\Sigma,\vectorize{\symmeasure})$, since $\vectorize{\symmeasure}$ is a vector of finite measures, we can define a new probability measure $\symmeasure_\featureset$ such that
\begin{equation*}
	\symmeasure_\featureset(Q)= \frac{\sum_{i=1}^n \symmeasure_i(Q)}{\sum_{i=1}^n \symmeasure_i(\featureset )}=\sum_{i=1}^n\frac{\symmeasure_i(Q)}{\Vert \vectorize{\symmeasure}(\featureset )\Vert_1}
\end{equation*}
Every normalized measure $\symmeasure_i/\Vert \vectorize{\symmeasure}(\featureset )\Vert_1$ is absolutely continuous with respect to $\symmeasure_\featureset$; hence, by the Radon-Nikodym theorem, there must be functions $\prob(\labely = i\conditional \individual)\colon \featureset\to [0,+\infty]$ such that
\begin{equation*}\label{radon-nikodym}
	\frac{\symmeasure_i(Q)}{\Vert \vectorize{\symmeasure}(\featureset )\Vert_1}= \int_Q \prob(\labely = i\conditional \individual) \symmeasure_\featureset(\diff \individual)
\end{equation*}
In particular, it's possible to choose them such that $\sum_{i=1}^n \prob(\labely = i\conditional \individual) =1$ for all $\individual\in \featureset$. Moreover, in this case $\symmeasure_\featureset$ remains an atomless probability measure, since the measures $\symmeasure_i$ are already atomless. So, as proven in \cite{dvoretzky1951relations}, we have that any slicing of $\featureset $ is equivalent to a decision function $\model\colon \featureset\to \labelset$ and vice versa, without the need of extra dimensions as in Theorem~\ref{teo_extra_dim}.
\subsection{The Simplex as the One Cake}
\label{subsec:the one cake}
Being able to derive probabilities $\prob(\labely \conditional \individual)$ for any cake-cutting problem provides a different point of view on how to approach the problem: $\prob(\labely \conditional \individual)$ can not only be seen as a conditional probability but also as a measurable function $\prob(\labely \conditional \individual)\colon \featureset\to \Delta^n $.

This means that we can enrich the simplex $\Delta^n$ with the push-forward measure defined by $\symmeasure_\Delta:=\symmeasure_\featureset\circ \prob(\labely \conditional -)^{-1}$.
Moreover, we can also define new conditional probability distributions $\prob_\Delta(\labely \conditional -)$ on $\Delta^n$ such that
\begin{equation*}
	\prob_\Delta(\labely = i\conditional \vectorize{t}) := t_i\text{ for all $\vectorize{t}=(t_1,\dots,t_n)\in \Delta^n$}
\end{equation*}
meaning that the conditional distributions on $\Delta^n$ correspond to the identity function.

The probability measure $\symmeasure_\Delta$, together with the conditional distributions $\prob_\Delta(\labely \conditional \vectorize{t})$, allows us to discuss probabilistic models and, by extension, cake-cutting on $\Delta^n$.

In particular, if we consider a probabilistic decision function $\model_{\Delta}\colon \Delta^n\to \labelset$, we have
\begin{align*}
	&\prob(\model_{\Delta} = j,\labely = i) 
	= 
	\\ 
	&\int_{\Delta^n} \prob(\model_{\Delta}(\vectorize{t})=j) \prob_\Delta(\labely = i\conditional \vectorize{t}) \symmeasure_\Delta(\diff \vectorize{t})
	= \\ 
	&\int_{\Delta^n} \prob(\model_{\Delta}(\vectorize{t})=j)\cdot t_i\cdot\symmeasure_\Delta(\diff \vectorize{t})
	=\\
	&\int_{\featureset}
	\prob\left(\model_{\Delta}\left(\prob\left(\labely =j \conditional \individual\right)\right)\right) \prob(\labely = i\conditional \individual) \symmeasure_\featureset(\diff \individual)
	=\\
	&\prob(\model_\Delta\circ \prob(\labely \conditional -)=j, \labely = i)
\end{align*}
So, the confusion matrix of any model on $\featureset $ that only depends on the conditional probabilities can be seen as the matrix for a decision on $\Delta^n$, and vice versa. But more importantly, as a consequence of Weller's theorem \cite{weller1985fair} stated in the next subsection, \emph{any} model $\model$ on $\featureset $ is equivalent to a decision $\model_{\Delta}$ that only depends on $\prob(\labely \conditional \individual)$. Which means that $\Delta^n$, together with conditional probability $\prob_\Delta(\labely \conditional \vectorize{t}):= \vectorize{t}$, provides a universal space for classification problems where the only variable that changes is the underlying probability $\symmeasure_{\Delta}$.

Critically, we must be aware that the push-forward measure $\symmeasure_{\Delta}$ may have atoms, even if $\symmeasure_\featureset$ does not. This is also often practically true in reality, for tabular data especially. If the attributes collected in $\featureset $ are discrete, then even if we assume an infinite population size granting an atomless measure $\symmeasure_\featureset$, the push-forward measure $\symmeasure_{\Delta}$ will still have atoms. This is because the possible combination of attributes remains finite, and hence the probability measure on $\Delta^n$ will be discrete. 

This explains the need of Theorem~\ref{teo_extra_dim} to maintain the connection with cake-cutting, since it allows us to say that the space $\Delta^n\times [0,1)$ provides a universal cake where the players $\symmeasure_i$ are defined by
\begin{equation*}
	\symmeasure_i(Q\times [a,b]) := 
	(b-a)\cdot
	{\int_Q t_i\cdot \symmeasure_{\Delta}(\diff \vectorize{t})}
\end{equation*}
for some probability measure $\symmeasure_\Delta$ on $\Delta^n$. 
\subsection{Cake-Cutting Results for Classification}\label{subsec-cake-class}
Let's now return for a moment to the definition of Pareto-optimality and discuss in more detail some important differences. Since we now know that cake-cutting and classification are two sides of the same coin, it seems natural to apply the Pareto pre-order defined on slicings as a pre-order on models as well. This is indeed how we will approach the problem, yet it is not the most general way to define an order on predictive models.

What we consider a "good model" in classification is one that minimizes errors. 
Hence, in full generality, given two models $\model,\model'$, we can say that one is better $\model\paretoless_e \model'$ than the other according to errors by defining $\paretoless_e$ such that
\begin{equation*}
	\model\paretoless_e \model' \iff \prob(\model= j, \labely = i)\geq \prob(\model'= j, \labely = i)\text{ for all }i\neq j
\end{equation*}
However, the Pareto pre-order on models $\paretoless$ is defined as
\begin{equation*}
	\model\paretoless \model' \iff \prob(\model= i, \labely = i)\leq \prob(\model'= i, \labely = i)\text{ for all }i\in [n]
\end{equation*}
If $n>2$, these two definitions are \emph{not} equivalent. In particular, it does hold that $\model\paretoless_e \model'\Rightarrow \model\paretoless \model'$, but unless we are discussing binary classification, the converse $\model\paretoless \model'\Rightarrow \model\paretoless_e \model'$ is \emph{false}.
\medskip
\begin{example}
	You have been tasked with building an AI model to predict blood pressure without the need for direct measurements. There are three possible labels, $\labelset=\{low, even, high\}$, and you have developed two algorithms, $\model,\model'\colon \featureset\to \labelset$, with the following confusion matrices
	\begin{equation*}
		\begin{array}{cc|ccc}
			\multicolumn{2}{c|}{\multirow{2}{*}{$M(\model)$}} 
			& &\model= &\\ 
			 && low & even & high\\
			\hline
			&low & 12\% & 15\% & 3\%\\
			\labely =&even & 15\% & 15\% & 0\% \\
			&high & 0 & 0 & 40\%
		\end{array}
		\hspace*{1cm}
		\begin{array}{cc|ccc}
			\multicolumn{2}{c|}{\multirow{2}{*}{$M(\model')$}} 
			& &\model= &\\ 
			 && low & even & high\\
			\hline
			&low & 15\% & 5\% & 10\%\\
			\labely =&even & 15\% & 15\% & 0\% \\
			&high & 0 & 0 & 40\%
		\end{array}
	\end{equation*}
	Looking at the diagonals, we can tell that $\model\paretostrictless \model'$, but the error rates cannot be compared directly. This means we cannot claim that $\model\paretoless_e \model'$, nor $\model\paretomore_e \model'$. In particular, it's debatable whether $\model$ is better than $\model'$, since predicting high blood pressure in people with low blood pressure can be more serious than predicting even blood pressure instead. 
\end{example}
\medskip
That said, by limiting ourselves to the use of $\paretoless$ only, we naturally inherit a lot of useful properties and results from cake cutting.

Similarly to what we have done for the Pareto pre-order, we can redefine concepts from cake-cutting as concepts for classification. For instance, $\ips(\featureset )$ can be considered as the set of diagonals of all possible confusion matrices for a classification problem. 

We dedicate the rest of this section to introduce a series of well-known results from cake-cutting theory restated using classification terminology. We start with one of the major results from \cite{dvoretzky1951relations} which generalizes a theorem from Lyapunov \cite{liapounoff1940fonctions}:
\medskip
\begin{theorem}[Generalized Lyapunov's Theorem \cite{dvoretzky1951relations}]
\label{teo_Dvoretsky}
Consider a classification problem on $\featureset $.
The set of achievable confusion matrices
\begin{equation*}
	\{M(\model)\colon \model\text{ is a model on }\featureset \}
\end{equation*}
is closed and convex. In particular, so is $\ips(\featureset )$.
\end{theorem}
\medskip
This finds an important use in the following well-established cake-cutting theorem, which states that unless there exist data points where certain labels can be excluded with absolute certainty, suboptimal models can always be improved in any direction. This is tied to the concept of aleatoric uncertainty, defined as follows:
\medskip
\begin{definition}\label{def_uncertainty}
	A classification problem on $\featureset$ is said to have \emph{aleatoric uncertainty} if
	\begin{equation*}
		\symmeasure_\featureset\Big(\{\individual\in \featureset \colon \prob(\labely = i\conditional \individual)=0\text{ for some }i\in[n] \}\Big) = 0
	\end{equation*}
	That is, for almost all $\individual\in \featureset$, no label can be excluded with absolute certainty. 
\end{definition}
\medskip
\begin{theorem}\label{teo_IPS_open}\Copy{teo_IPS_open}{
	Let $\model\colon \featureset\to \labelset$ be a model on a classification problem $\featureset$ with aleatoric uncertainty.
	The model $\model$ is \emph{not} Pareto-optimal if and only if there exists $\varepsilon>0$ such that
	\begin{equation*}
	\left\{\diag M( \model)\right\} +[0,\varepsilon]^n\subseteq \ips(\featureset )
	\end{equation*}}
	\begin{proof}
		The proof is equivalent to the cake-cutting version that can be found in \cite{barbanel_taylor} for Lemma 1.13, by combining it with Theorem~\ref{teo_Dvoretsky}. We provide it in the \proofref{teo_IPS_open} for the classification case. 
	\end{proof}
\end{theorem}
Another important result is Weller's theorem \cite{weller1985fair}, which provides a sufficient and necessary condition to construct Pareto-optimal models. 
\begin{figure}[t!]
	\centering
	\begin{subfigure}[t]{.32\textwidth}
		\centering
		\includegraphics[width=\textwidth]{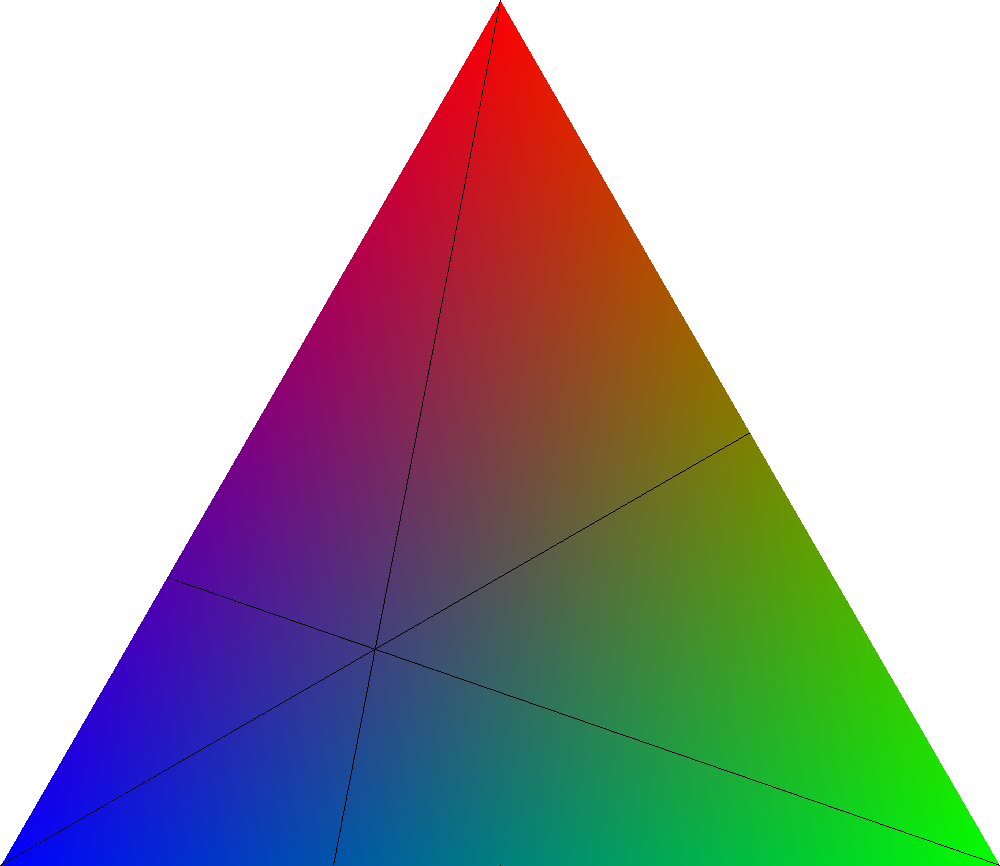}
		\caption
		{
			Lines intersect at point $(0.25,0.25,0.5)$
		}
	\end{subfigure}%
	\hfill
	\begin{subfigure}[t]{.32\textwidth}
		\centering
		\includegraphics[width=\textwidth]{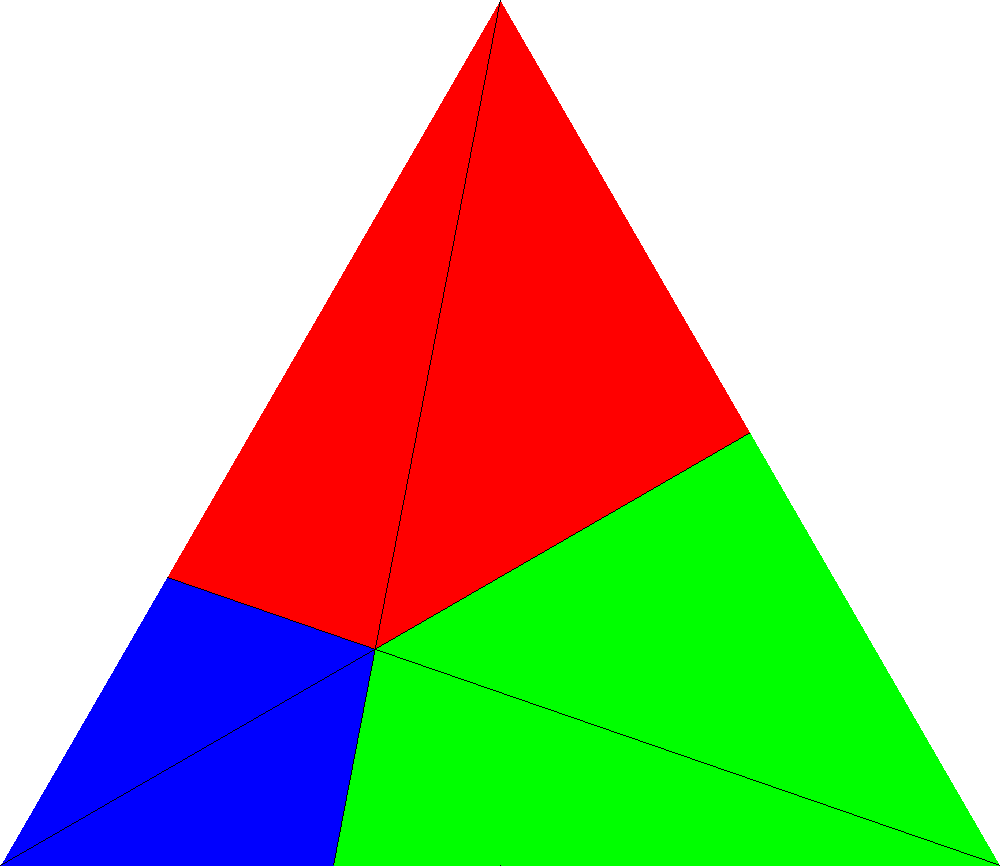}
		\caption
		{
			$\vectorize{{w}}^{(k)}=\left(\dfrac{1}{4},\dfrac{1}{4}, \dfrac{1}{2}\right)$
		}
	\end{subfigure}
	\hfill
	\begin{subfigure}[t]{.32\textwidth}
		\centering
		\includegraphics[width=\textwidth]{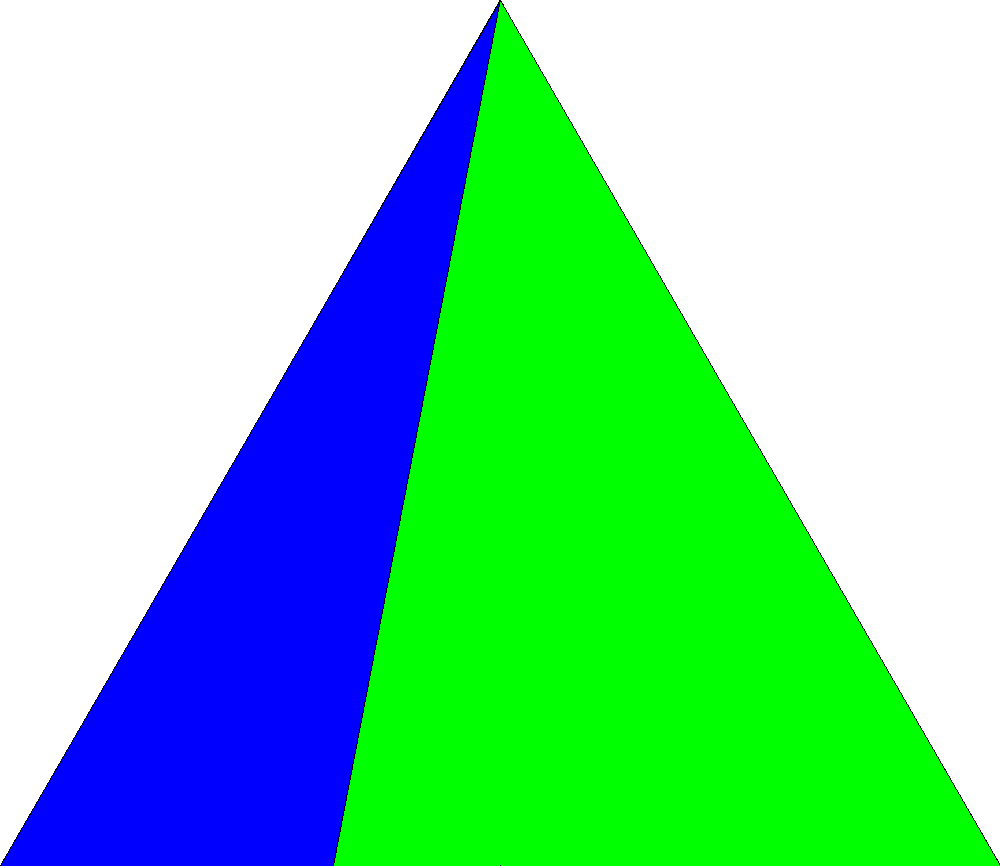}
		\caption
		{
			$\vectorize{{w}}^{(k)}=\dfrac{1}{k}\cdot\left(k-\dfrac{1}{2},\dfrac{1}{6}, \dfrac{1}{3}\right)$
		}
	\end{subfigure}
	\caption{Optimal decisions according to Weller's theorem on the $\Delta^3$ simplex with labels $\labelset=\{\text{red},\text{green}, \text{blue} \}$. The RGB decomposition of a color represents the conditional distribution $\prob(\labely \conditional \individual)$.}\label{fig_weller}
\end{figure}
\medskip
\begin{theorem}[Weller's Theorem \cite{weller1985fair}]\label{teo_weller}
	A model $\model\colon \featureset \to \labelset$ is Pareto-optimal if and only if there exists a sequence $(\vectorize{{w}}^{(k)})_{k\in\NN}$ in $\Delta^n\cap (0,1)^n$ such that: 
	\begin{itemize}
		\item For all $i,j\in [n]$, the limit $\lim_{k\to \infty} {{w}_i^{(k)}}/{{w}_j^{(k)}}$ converges in $[0,\infty]$.
		\item For all $i\in[n]$, almost everywhere on $\featureset $,
		\begin{equation*}
		\prob(\model(\individual)=i)>0 \Rightarrow \exists\ \overline{k}\in\NN\text{ s.t. }\frac{\prob(\labely = i\conditional \individual)}{\prob(\labely = j\conditional \individual)}\geq \sup_{k\geq \overline{k}}\left(\frac{{w}_i^{(k)}}{{w}_j^{(k)}}\right) \ \forall j\in [n]
		\end{equation*}
	\end{itemize}
	Such a sequence $(\vectorize{{w}}^{(k)})_{k\in\NN}$ is said to be $w$-associated with $\model(\individual)$.
\end{theorem}
\medskip
Figure \ref{fig_weller} provides a visual example for Weller's theorem. For the binary case this theorem can be restated as follows:
\medskip
\begin{corollary}\label{cor_binary_weller}\Copy{cor_binary_weller}{
	A binary model $\model\colon \featureset\to \binaryset$ is Pareto-optimal if and only if there exists $t\in [0,1]$ such that almost everywhere on $\featureset $, it holds
	\begin{equation*}
		\prob(\model(\individual)=1)=
		\begin{cases}
			0 &\text{if }\ \prob(\labely = 1\conditional \individual) < t\\
			1 &\text{if }\ \prob(\labely = 1\conditional \individual) > t\\
			t &\text{if }\ t\in\{0,1\} \text{ and } \prob(\labely = 1\conditional \individual) = t\\
			q(\individual) &\text{if }\ t\in(0,1) \text{ and }\prob(\labely = 1\conditional \individual) = t
		\end{cases}
	\end{equation*}
	where $q\colon \featureset\to [0,1]$ is some measurable function.}
	\begin{proof}
		Direct application of Weller's theorem. An in-depth proof can be found in the \proofref{cor_binary_weller}.
	\end{proof}
\end{corollary}
\begin{figure}[t!]
	\centering
	\begin{subfigure}[t]{.5\textwidth}
		\centering
		\includegraphics[width=.8\textwidth]{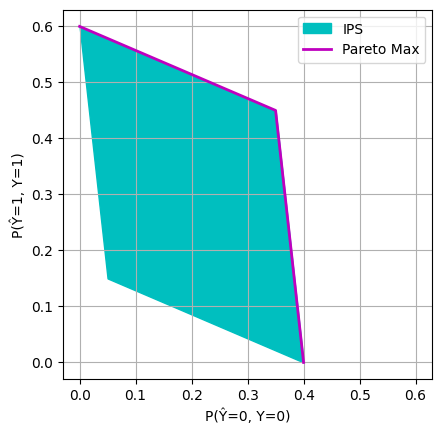}
		\caption
		{
			$\ips$ and Pareto-optimal models.
		}
	\end{subfigure}%
	\hfill
	\begin{subfigure}[t]{.5\textwidth}
		\centering
		\includegraphics[width=.8\textwidth]{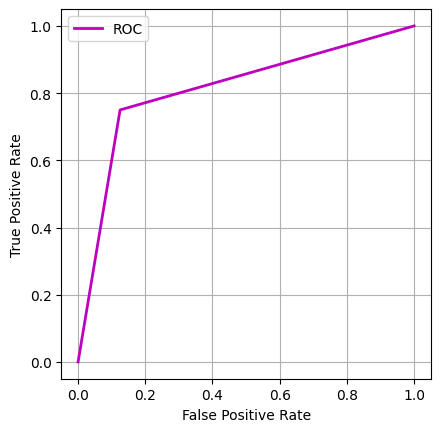}
		\caption
		{
			The $\roc$ curve for the same problem.
		}
	\end{subfigure}
	\caption
	{
		$\ips$ and $\roc$ curve for a binary classification problem where half the population has $\prob(\labely = 1\conditional \individual)=0.9$ and the other half $\prob(\labely = 1\conditional \individual)=0.3$. Straight edges correspond to atoms of the push-forward measure $\symmeasure_{\Delta}$.
	}\label{fig_ROC}
\end{figure}

The corollary shows that, for the binary case, optimal solutions are achieved by selecting a threshold for the probability distribution.
For non-binary classification, a sequence $\vectorize{{w}}^{(k)}$ can be seen as the generalization of what a threshold represents in the binary case.

The corollary formally justifies what is already intuitively done in binary classification with $\roc$ curves. In fact, there is a direct connection between the $\roc$ curve and the $\ips$. The $\roc$ curve is the curve traced by $(\prob(\model=1\mid\labely = 0), \prob(\model=1\mid\labely = 1)) \in [0,1]^2$ as the decision threshold changes, using the predicted scores given by a model as a proxy for $\prob(\labely = 1 \mid \individual)$. Thanks to Weller's theorem, we know this corresponds to the Pareto-optimal models for the problem, if the predicted scores are correct. It is easy to observe that the best $\roc$ curve and the optimal points of the $\ips$ are identical, merely mirrored and rescaled, as shown in Figure \ref{fig_ROC}.

Yet, the formal approach of Corollary~\ref{cor_binary_weller} highlights an important problem that the more intuitive approach fails to notice. In general, there exist points on the $\roc$ curve that cannot be obtained with a deterministic threshold.

Suppose there is a set $Q \subset \featureset$ of positive measure such that $\prob(\labely = 1\mid \individual) = t$ is constant on $Q$. If the threshold chosen is exactly $t$, then some optimal decisions can only be obtained by partitioning the set $Q$ or using a probabilistic model. This means that either we need to make different decisions on data points having the same conditional probability, or we need to assign the label randomly. This can also be characterized visually, as expressed in the following lemma:
\medskip
\begin{lemma}\label{lemma_segment}
	For a binary classification problem, the push-forward measure $\symmeasure_{\Delta}$ is not atomless if and only if $\ips(\featureset)$ has straight edges.
	Moreover, if a Pareto-optimal model $\model$ corresponds to a point in the interior of a straight edge of $\ips(\featureset )$, then either $\model$ is not deterministic or does not solely depend on $\prob(\labely \conditional \individual)$.
	\begin{proof}
		The proof simply combines Corollary~\ref{cor_binary_weller} with the explanations provided in \cite{barbanel_taylor} from Chapter 12 and Theorem 2.6.
	\end{proof}
\end{lemma}
\section{The Fairness Optimization Problem}\label{sec_fairness_optimization}
In this section, we formally introduce the fairness optimization problem and lay down the definitions needed to continue with our discussion. We also provide some characterizations of the proposed definitions, which make use of the concepts introduced in the previous sections.

As mentioned in Section~\ref{sec_understanding_cherry}, the study of group fairness depends on demographic subgroups of the population, called \emph{sensitive groups}, which partition the set $\featureset$. To simplify the discussion, we assume the existence of only two groups, $\{\agroup{0},\agroup{1}\}$, but all the results can be easily generalized to more groups and are independent of this assumption. Moreover, the use of "$\agroup{0}$" and "$\agroup{1}$" is only meant as an indication for the sensitive groups; it does not have semantic meaning. Their use is simply syntactic sugar for the reader, reminding them of the sensitive connotation. With a bit of notation abuse, we often write "$A=\agroup{0}$" to indicate the set $\agroup{0}$, and similarly for $\agroup{1}$. Every sensitive group naturally inherits all properties of the classification problem from $\featureset$ via conditioning.

In our discussion of fairness, we will focus solely on binary classification problems $(\labelset=\binaryset)$. This avoids the shortcomings mentioned at the start of Section~\ref{subsec-cake-class}, which would present a challenge otherwise.

A fairness problem is divided into two pieces: an evaluation metric that measures the predictive quality of a model, and a fairness measure that quantifies the fairness of a model. In order to define these two concepts, we need to introduce the notion of rate-dependent functions, which are functions that output a value when given a model.
\medskip
\begin{definition}[Rate-dependent function] 
	We say that a function $\REval{\model}{}$ is \emph{rate-dependent} if: 
	\begin{enumerate} 
		\item\label{point_one} 
		{There exists a function $\realratedependent{E}{\varrate{0},\varrate{1}}\colon [0,1]^2 \times \mathcal{B}([0,1]^2) \to \RR$ such that 
		\begin{equation*} 
			\REval{\model}{} = \realratedependent{\ips(\featureset)}{\jointrate{0}{\model}{}, \jointrate{1}{\model}{}} 
		\end{equation*} 
		for all models $\model\colon \featureset \to \binaryset$.} 
		\item The function $\realratedependent{E}{\varrate{0},\varrate{1}}$ is continuous, where continuity for the Borel $\sigma$-algebra $\mathcal{B}([0,1]^2)$ is understood with respect to the Fréchet-Nikodym metric $\lambda\left(E \triangle E'\right)$. That is, 
		\begin{equation*} 
			\lambda\left(E \triangle E'\right) := \lambda(E' \smallsetminus E) + \lambda(E \smallsetminus E') 
		\end{equation*} where $\lambda$ is the Lebesgue measure on $[0,1]^2$. 
		\item The function $\realratedependent{E}{\varrate{0},\varrate{1}}$ is continuously differentiable in $(\varrate{0}, \varrate{1})$. 
	\end{enumerate} 
	We will write $\symratedependent(\varrate{0}, \varrate{1}, E)$ instead of $\realratedependent{E}{\varrate{0}, \varrate{1}}$, since it is clear from the variables used which function is intended. 
\end{definition}
\medskip
Essentially, we want rate-dependent functions to be functions that depend on the confusion matrix, more specifically on the joint negative rate $\varrate{0} = \jointrate{0}{\model}{}$ and the joint positive rate $\varrate{1} = \jointrate{1}{\model}{}$ of the model. Yet, this alone is not enough to make rate-dependent functions useful for our purposes. There is a problem if we require only this property: models on different classification problems could have the same confusion matrix, but the underlying problems could be extremely different.

If we want to use rate-dependent functions to evaluate models, it would be reasonable to expect that the evaluation of a model might depend on how difficult the problem is. For some problems, blindly assigning a label might be close to the best decision, while for others, it might be far from it. The rate-dependency definition captures this property, which is expressed in the first point by requiring that a rate-dependent function also depends on the $\ips$.

On the other hand, without specifying some form of continuity, we could end up with ill-behaved functions that change abruptly when we slightly change the classification problem. So, we would like to consider only functions that behave as expected, as expressed by the second point, which requires that the rate-dependent function continuously depends on $ \lambda\left(\ips(\featureset) \triangle \ips(\featureset')\right) $ for two classification problems $\featureset, \featureset'$. 

In general, the symmetric difference is just a pseudo-metric, but for $\ips$s specifically, it is actually a well-defined distance. By extension, it is possible to talk about the distance between two classification problems by considering the distance between their $\ips$s. In general, the larger the $\ips$, the easier the problem.

The fact that $\REval{\model}{}$ depends on the $\ips$ also implies that the output of a rate-dependent function is invariant under the permutation of the conditional distributions. More specifically, if we consider a model $\model$ that depends only on the conditional distribution $\prob(\labely \mid \individual)$, then the value $\REval{\model}{}$ does not change for different classification problems, as long as the fraction of individuals with a given conditional distribution stays the same. This is important because it ensures that individuals are fully described by their conditional distribution, and no person is more important than another. This is formally proved in the following lemma:
\medskip
\begin{lemma}\label{lemma_eloss_exists}\Copy{lemma_eloss_exists}{
	Consider a function $\REval{\model}{}$, the following are equivalent:
	\begin{enumerate}
		\item $\REval{\model}{}$ satisfies Point \ref{point_one} of the definition of rate-dependent functions.
		\item If the push-forward measure on $\Delta^2$ induced by two classification problems $\featureset,\featureset'$ is the same, then for all models $\model\colon\featureset\to\binaryset$ and $\model'\colon\featureset'\to\binaryset$ it holds:
		\begin{itemize}
			\item[{}]If $\varphi\colon [0,1]\to\binaryset$ is a measurable function s.t. 
			$\model\paretoeq\varphi\circ \prob_{\featureset}(\labely=1 \conditional -)$ and $\model'\paretoeq\varphi\circ \prob_{\featureset'}(\labely=1 \conditional -)$, then $\REval{\model}{}= \REval{\model'}{}$
		\end{itemize}
	\end{enumerate}
	}
	\begin{proof}
		In the \proofref{lemma_eloss_exists}.
	\end{proof}
\end{lemma}
\medskip
This last lemma also shows another way to characterize rate-dependent functions: these are continuous functions that fully depend on the induced cake-cutting problem described in Section~\ref{subsec:the one cake}. Essentially, the lemma states that if two models are equivalent when seen as models on the simplex, then a rate-dependent function will treat them as the same model.

This alternative definition shows how broad the class of rate-dependent functions is, which the original definition doesn't intuitively capture. Requiring the direct dependency on the $\ips$ in the outlined definition seems more restrictive than making the function dependent on the induced cake-cutting problem on the simplex, since, as we explained, \emph{any} cake-cutting problem is a problem on the simplex. However, this is not true, since there is a one-to-one correspondence between the $\ips$ for a problem and its induced push-forward measure on $\Delta^2$.

Rate-dependent functions provide us with the tools to define evaluation metrics and fairness measures.
\medskip
\begin{definition}[Evaluation Metric]
	A rate-dependent function $\EEval{\model}{}$ is said to be an \emph{evaluation metric} on $\featureset$, if it is monotonic according to the Pareto order. That is,
		\begin{equation*}
			\model\paretoless \model'
			\Rightarrow
			\EEval{\model}{}\leq
			\EEval{\model'}{}
		\end{equation*}
		for all models $\model,\model'\colon \featureset\to \binaryset$.
		Moreover, we say that $\EEval{}{}$ is \emph{strict} if the property holds with strict inequalities.
\end{definition}
\medskip
\begin{definition}[Delta Fairness Measure]
	Given sensitive groups $\{\agroup{0},\agroup{1}\}$, a \emph{delta fairness measure} is a function of the form
	\begin{equation*}
		 \left\vert\FEval{\model\restricted{\agroup{0}}}{} - \FEval{\model\restricted{\agroup{1}}}{}\right\vert
	\end{equation*}
	where $\FEval{}{}$ is a rate-dependent function called the \emph{(fairness) utility function}.
\end{definition}
\medskip
\begin{definition}[Fairness Problem]
	A \emph{(group) fairness problem} on $\featureset$ is an instance of the following problem:
	\begin{equation*}
		\FairnessProblem{}
	\end{equation*}
	where $\EEval{}{}$ is an evaluation metric and $\FEval{}{}$ a utility function for a delta fairness measure. A model $\model$ is called a \emph{solution} to the fairness problem if $\model$ maximizes the objective.
\end{definition}
\medskip
The definition of an evaluation metric is quite natural: essentially, it states that an evaluation metric is a function that outputs a higher value (or more precisely non-lower value) when a model improves. 
If an evaluation is also strict, any Pareto-superior model will have a strictly higher evaluation.

Instead, a (delta) fairness measure is not an evaluation metric but depends on the absolute difference of a utility function $\FEval{}{}$ defined on each sensitive group. The only condition $\FEval{}{}$ has to satisfy is to be rate-dependent, which means that fairness measures depend solely on the errors of the model within each sensitive group and can be computed by examining the confusion matrices for the two groups.

Notice that this has two important implications. First, two models that are Pareto-equivalent on $\featureset$ may have different fairness measurements because they could perform differently across each sensitive group. The other implication is that this definition of fairness only concerns what is called \emph{group} fairness, since the use of a rate-dependent function $\FEval{}{}$ guarantees no preference between individuals from the same sensitive group having the same conditional probability distribution.

Also, our definition of fairness is symmetric for each sensitive group, which often could not be the case in practice. In general, fairness is often meant to protect minorities suffering from discrimination, which could require non-symmetric objectives for each considered group. Our theory can be extended to include this case, but for simplicity, we will focus on the delta-fairness case.
\medskip
\begin{example}[Precision]
	Precision is a strict evaluation metric, since it can be rewritten as:
	\begin{equation*}
		\prob(\labely = 1\conditional \model=1)=\frac{\jointrate{1}{\model}{}}{\jointrate{1}{\model}{}+\prob(\labely =0)-\jointrate{0}{\model}{}}
	\end{equation*}
	The value $\prob(\labely = 0)$ depends on the classification problem on $\featureset $, and it is continuous with respect to the measure of $\ips(\featureset )$.
\end{example}
\medskip
\begin{example}[Loss Functions]
	Consider the 0-1 loss function $\mathcal{L}(\modeloutcome, \labeloutcome ) := \mathbbm 1_{\modeloutcome\neq \labeloutcome }$.\ The expected loss is \begin{equation*} \EEsub_{\substack{ \individual\sim \symmeasure_\featureset\\ \labeloutcome \sim \prob(\labely \conditional \individual)\\ \modeloutcome \sim \model(\individual) }}\mathcal{L}(\modeloutcome, \labeloutcome ) = 1 - \jointrate{1}{\model}{} - \jointrate{0}{\model}{} \end{equation*} In particular, the 0-1 loss is one minus a strict evaluation metric.
More generally, if we consider any loss function $\mathcal{L}(\modeloutcome, \labeloutcome )$ for multi-label classification, we can rewrite the previous expected loss as \begin{align*} \int_{\featureset}\bigg(\sum_{i,j\in[n]}\mathcal{L}(j, i)\prob({\model(\individual)= j})\prob(\labely=i\conditional \individual)\bigg)\symmeasure_\featureset(\diff \individual) = \sum_{i,j\in[n]}\mathcal{L}(j, i)\prob(\model = j, \labely = i) \end{align*} which shows that, in general, expected loss functions are functions of the confusion matrix $M(\model)$ when the predictions are sampled from $\model(\individual)$. Moreover, if we assume no penalty for correct predictions (i.e., $\mathcal{L}(j,i) = 0$ if $i = j$) and a constant penalty for errors on the same label (i.e., $\mathcal{L}(j,i) = \phi_i$ for $i \neq j$, which is always the case for binary classification), we have: \begin{equation*} \sum_{i,j\in[n]}\mathcal{L}(j, i)\prob(\model = j, \labely = i) = \sum_{i\in[n]}\phi_i\cdot\prob(\labely = i) - \phi_i\cdot\jointrate{i}{\model}{} \end{equation*} This means that the expected loss can be seen as a decreasing function depending on the joint rates $\jointrate{i}{\model}{}$ and $\ips(\featureset)$. For binary classification, using the terminology developed so far, we can say that the expected loss behaves like the opposite of an evaluation metric.
\end{example}
\medskip
\begin{example}[Immediate Utility]
For all $t\in [0,1]$, Immediate Utility ${U}_t$ is an evaluation metric, since it can be rewritten as:
\begin{equation*}
	{U}_t(\model) = (1-t)\cdot\jointrate{1}{\model}{}+t\cdot\jointrate{0}{\model}{} -t\cdot\prob(\labely = 0)
\end{equation*}
which is non-decreasing in $\jointrate{0}{\model}{}$ and $\jointrate{1}{\model}{}$. Notice that if $t\in\binaryset$, then ${U}_t$ is not strict.
\end{example}

\section{When Optimizing Fairness Leads to Cherry-Picking}\label{sec_from_fairness_to_cherry}
We can now find a proper characterization of the cherry-picking definition we introduced in Section~\ref{sec_understanding_cherry}.
\medskip
\begin{theorem}\label{teo_atomic_cherry_picking}
	A model $\model$ almost never cherry-picks if and only if, for each sensitive group $A \in \{\agroup{0}, \agroup{1}\}$ and for all $\vectorize{v}\in (0,1)^2$, it holds that:
	\[
			\diag M(\model\restricted{A}) +\vectorize{v}\not\in\ips(A)
	\]
	In particular, if a model $\model$ is Pareto-optimal, then it almost never cherry-picks. Moreover, if $X$ has aleatoric uncertainty, the converse is also true.
	\begin{proof}
		Formally a model almost-never cherry-picks if it is equal to a non-cherry-picking model up to a set of measure zero. A model does not cherry-pick if and only if it holds for each $\individual\in A$
		\begin{equation*}
			\prob(\model(\individual)=1)>0\Rightarrow \model(\individual')=1\text{ for all }\individual'\in A\text{ s.t. }\prob(\labely =1\conditional \individual')>\prob(\labely =1\conditional \individual)
		\end{equation*}
		where $A$ is a sensitive group. This is equivalent to saying that there exists $t\in [0,1]$ such that
		\begin{equation*}
			\prob(\model\restricted{A}(\individual)=1) = 
			\begin{cases}
				1 &\text{if } \prob(\labely =1\conditional \individual)>t \\
				0 &\text{if } \prob(\labely =1\conditional \individual)<t \\
				q(\individual) &\text{if } \prob(\labely =1\conditional \individual) = t
			\end{cases}
		\end{equation*}
		This is always Pareto-optimal if $\featureset$ has aleatoric uncertainty or $t\in(0,1)$, since the set of $\individual$ such that $\prob(\labely =1\conditional \individual)\in \binaryset$ has measure zero. So the previous model is almost-everywhere equal to a model satisfying the condition for Pareto-optimality given in Corollary~\ref{cor_binary_weller}. If $\featureset$ does not have uncertainty and $t\in\{0,1\}$, then $\prob(\model\restricted{A}(\individual)=1-t)$ is already the highest possible, so it cannot be improved.
	\end{proof}
\end{theorem}
Notice that, as we discussed after Corollary~\ref{cor_binary_weller}, if the set of $\individual \in A$ such that $\prob(\labely = 1 \mid \individual) = t$ has positive measure, it might be necessary to take different decisions for people having the same conditional probability. This can be seen as a stronger form of cherry-picking, where the condition for cherry-picking holds not just for points with strictly lower conditional probabilities, but also for points with the same conditional probability.

Yet, for this stronger version of cherry-picking, guaranteeing non-cherry-picking models for \emph{any} fairness measure is impossible. As shown in Lemma~\ref{lemma_segment}, if the push-forward measure $\symmeasure_{\Delta}$ is not atomless, then the $\ips$ has straight edges, and any point on the interior of a straight edge would cherry-pick according to this stronger version. The classification problem on $\Delta^2$ often has atoms, even when $\symmeasure_\featureset$ is atomless, as discussed in Section~\ref{subsec:the one cake}.

Technically, a model that is Pareto-optimal on each sensitive attribute might still cherry-pick, but Theorem~\ref{teo_atomic_cherry_picking} guarantees that we can correct it by changing the model on a set of measure zero.

This provides us with a tool to understand under which conditions non-cherry-picking models exist. We start by generalizing the results from \cite{corbett2017algorithmic, menon2018cost}, showing that the fairness measures they discuss always allow for non-cherry-picking models.

More generally, the following theorem shows a general condition that guarantees when a fairness measure behaves well under optimization.
\medskip
\begin{theorem}\label{teo_when_works}\Copy{teo_when_works}{
	Consider a utility function $\FEval{}{}$ such that ${\partial_{\varrate{0}}{\FEval{}{}}}\cdot
	{\partial_{\varrate{1}}{\FEval{}{}}}
	\leq 0$. Then, any fairness problem admits non-cherry-picking solutions.\\
	Formally, for any evaluation metric $\EEval{}{}$ and sensitive groups $\{\agroup{0},\agroup{1}\}$, there exists a model
	\begin{equation*}
		\optimalmodel \in \FairnessProblem{}
	\end{equation*}
	that does not cherry-pick.
	}
	\begin{proof}
		In the \proofref{teo_when_works}.
	\end{proof}
\end{theorem}
It's now easy to show that the fairness measures considered by \cite{corbett2017algorithmic,menon2018cost} satisfy the theorem's claim.
\medskip
\begin{theorem}\label{teo_dpd_eo_ect}
	Let $\FEval{}{}$ be a utility function defining one of the following delta fairness measures:
	\begin{itemize}
		\item (Conditional) Demographic Parity Difference:\\
		$\left\vert \prob( \model=1\conditional A=\agroup{1})-\prob(\model=1\conditional A=\agroup{0}) \right\vert $
		\item (Conditional) Equal Opportunity Difference:\\
		$\left\vert \prob( \model=1\conditional\labely = 1,A=\agroup{1})-\prob(\model=1\conditional\labely = 1,A=\agroup{0}) \right\vert $
		\item (Conditional) Equal Risk Difference:\\
		$\left\vert \prob( \model=0\conditional\labely = 0,A=\agroup{1})-\prob(\model=0\conditional\labely = 0,A=\agroup{0})\right\vert $
	\end{itemize}
	then any fairness problem always admits non-cherry-picking solutions.\\
	Formally, for any evaluation metric $\EEval{}{}$ and sensitive groups $\{\agroup{0},\agroup{1}\}$, there exists a model
	\begin{equation*}
	\optimalmodel \in
	\FairnessProblem{}
	\end{equation*}
	that does not cherry-pick.
	\begin{proof}
		Let's formally define the utility function $\FEval{\varrate{0},\varrate{1}}{E}$ for all the measures considered, and the corresponding gradients: 
		\begin{itemize}
			\item Demographic Parity Difference:
			\begin{equation*}
				\varrate{1}-\varrate{0}+\prob(\labely = 0\conditional A)\Rightarrow \partial_{\varrate{0}}\FEval{}{} = -\partial_{\varrate{1}}\FEval{}{}
			\end{equation*}
			\item Equal Opportunity Difference:
			\begin{equation*}
				\dfrac{\varrate{1}}{\prob( \labely = 1\conditional A)}\Rightarrow \partial_{\varrate{0}}\FEval{}{}  =0
			\end{equation*}
			\item Equal Risk Difference:
			\begin{equation*}
				\dfrac{\varrate{0}}{\prob(\labely = 0\conditional A)}\Rightarrow \partial_{\varrate{1}}\FEval{}{}=0
			\end{equation*}
		\end{itemize}
		In particular, regardless of which measure is taken, we can apply Theorem~\ref{teo_when_works} which proves the theorem.
	\end{proof}
\end{theorem}
\medskip
	This result has three major consequences:
	\begin{enumerate}
		\item As already mentioned previously, this shows that the results from \cite{corbett2017algorithmic,menon2018cost} can be generalized to any evaluation metric and not only apply to Immediate Utility. Therefore, the result can be seen as an improvement of the theoretical state-of-the-art.
		\item The result has practical consequences on how to find fair solutions for the three proposed measures. Since maximal solutions are Pareto-optimal for each sensitive group, to find a maximal solution, one only needs to identify suitable thresholds of the $\roc$ curves for each sensitive group. This approach is not new, and it has been used since the beginnings of the field of fairness \cite{massaging1,kamiran2012data, hardt2016equality}. Now, however, we also know that these methods not only discover fair solutions but can identify the best overall solution when the conditional probability for a data point is correctly modeled.
		\item As a consequence, we also find that for these specific fairness measures, there isn't an inherent conflict between fairness and cherry-picking. Moreover, Theorem~\ref{teo_when_works} extends this result to any fairness measure that satisfies its hypothesis.
	\end{enumerate}
	That said, these results for Demographic Parity, Equal Opportunity, and Equal Risk are far from being general. The condition proposed in Theorem~\ref{teo_when_works}, although general, is also somewhat counterintuitive since it requires that, regardless of the model we are considering, if we can find a better model for a sensitive group, then we can find one that is equally fair, if not more so.

However, depending on the context, this does not always align with our intuition for fairness: if a model is already performing extremely well for one sensitive group but not for another, it is reasonable to expect that a model which is even better for the first group should be considered less fair.

This type of group fairness is enforced when the utility function $\FEval{}{}$ of the considered fairness measure is an evaluation metric. Essentially, these fairness measures use an evaluation metric to compare the quality of the model for the two communities. Equal Opportunity and Equal Risk are two examples of such measures, as one compares the true positive rate and the other the false positive rate between the communities. Another well-known example is Predictive Parity (often called Calibration or Sufficiency), which compares the precision of the classifier between groups and is the focus of our initial example in Section~\ref{sec_understanding_cherry}.

In contrast, Demographic Parity does not use an evaluation metric for the utility function $\FEval{}{}$, since even if $\FEval{\model}{} = \prob(\model = 1 \mid A)$ is a rate-dependent function, it is not monotonic with respect to the Pareto order.

We are going to show that, for these evaluation-based fairness measures, the result of Theorem~\ref{teo_when_works} generally does not hold. More precisely, if the utility function $\FEval{}{}$ is a strict evaluation, then there exists a sensitive partition $\{\agroup{0},\agroup{1}\}$ and an evaluation metric $\EEval{}{}$ such that \emph{all} optimal models must cherry-pick. 
More formally:
	\begin{figure}[t!]
    \centering
    \begin{subfigure}[t]{.5\textwidth}
        \centering
        \includegraphics[width=.8\textwidth]{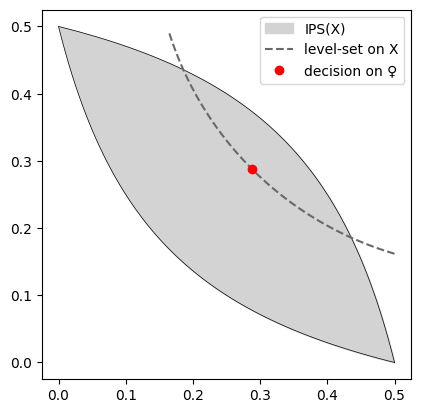}
        \caption
        {
            Point $(\varfix{0},\varfix{1})$ with level set for
            $
            \FEval{-}{}
            $.
        }
    \end{subfigure}%
    \hfill
    \begin{subfigure}[t]{.5\textwidth}
        \centering
        \includegraphics[width=.8\textwidth]{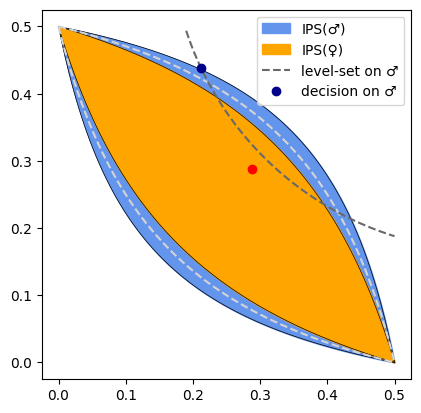}
        \caption
        {
            $(\varfix{0},\varfix{1})$ now corresponds to a model on $\agroup{0}$.
        }
    \end{subfigure}
    \caption
    {
        A visualization of the proof for Theorem~\ref{teo_bad_fairness}. After finding $(\varfix{0},\varfix{1})$ for a specific level set of $\FEval{}{}$, the set $\featureset $ is split into $\agroup{0}$ and $\agroup{1}$ and a new model is constructed based on the new level set $\FEval{}{}$ on $\agroup{1}$.
    }\label{fig_teo}
\end{figure}
	\medskip
	\begin{theorem}\label{teo_bad_fairness}
    Let the utility function $\FEval{}{}$ be a strict evaluation metric and $\featureset$ an atomless classification problem. If $\FEval{\model}{}$ is not constant for all Pareto-optimal models, then there exist sensitive groups $\{\agroup{0},\agroup{1}\}$ and evaluation metric $\EEval{}{}$ such that any solution
        \begin{equation*}
            \optimalmodel \in \FairnessProblem{}
        \end{equation*}
        must cherry-pick.    
        \begin{proof}
        Consider the interval $I$ defined as
        \begin{equation*}
            I:=\{\FEval{\model}{}\mid \model\colon X\to \binaryset \text{ is a non-trivial Pareto-optimal model}\}
        \end{equation*}
        we know it cannot be a singleton.
        Since $\FEval{}{}$ is continuous and the set of Pareto-optimal models generate a connected boundary on the $\ips$, it follows $I$ must indeed be an interval having open interior $(a,b)$ for some $a<b$.\\
        As a consequence, we have an open set $Q\subseteq \ips(\featureset)$ of points such that $\FEval{\varrate{0},\varrate{1}}{\ips(\featureset)}\in (a,b)$ for all $(\varrate{0},\varrate{1})\in Q$. In particular, there must be a point $(\varfix{0},\varfix{1})\in Q$ where
        \begin{equation*}
            \partial_{\varrate{0}}\FEval{\varfix{0}, \varfix{1}}{\ips(\featureset)}>0\text{ and }
            \partial_{\varrate{1}}\FEval{\varfix{0}, \varfix{1}}{\ips(\featureset)}>0
        \end{equation*}
        otherwise this would mean for all $(\varrate{0},\varrate{1})\in Q$
        \begin{equation*}
            \partial_{\varrate{0}}\FEval{\varrate{0}, \varrate{1}}{\ips(\featureset)}=0
            \text{ or }
            \partial_{\varrate{1}}\FEval{\varrate{0}, \varrate{1}}{\ips(\featureset)}=0
        \end{equation*}
        since $\FEval{}{}$ is an increasing differentiable function. We can prove that one of the two must always be equal to zero on $Q$, which is done in the Appendix in Theorem~\ref{teo_almost_egregium}.\\
        As a consequence, we have that $\FEval{}{}$ is constant in one direction on $Q$ which conflicts with the fact that $\FEval{}{}$ is strict. 
        Hence, the point $(\varfix{0},\varfix{1})$ must exist.\\
        Fix constants $0<\varepsilon,\varepsilon'<1$ and consider now a partition $\{\agroup{0},\agroup{1}\}$ of $\featureset $ constructed in the following way:
        \begin{itemize}
            \item $\ips({\agroup{0}})\subseteq \ips({\agroup{1}})$
            \item $\partial\ips({\agroup{0}})\cap \partial\ips({\agroup{1}})\subseteq  \partial \big([0,\prob(\labely = 0)]\times [0,\prob(\labely = 1)]\big)$ 
            \item $0<\lambda(\ips(\agroup{1})\smallsetminus \ips(\agroup{0}))<\varepsilon$
            \item $\symmeasure_\featureset(\agroup{0})=\varepsilon'$
        \end{itemize}
        where $\partial \ips$ represents the boundary of an $\ips$ and $\lambda$ is the Lebesgue measure.\\
        It's sufficiently intuitive that such a partition always exists, but a formal proof of this statement can be found in Corollary~\ref{cor_any_communities} in the Appendix.\\
        Moreover, if we set $\varepsilon$ small enough we get that 
        \begin{itemize}
            \item the point $(\varfix{0},\varfix{1})$ is in the interior of $\ips(\agroup{0})$,
            \item the partial derivatives of $\FEval{}{}$ are still strictly positive in a neighborhood of $(\varfix{0},\varfix{1})$,
            \item there exists a non-trivial Pareto-optimal model $\model_{\agroup{1}}\colon\agroup{1}\to\binaryset$ such that $\FEval{\model_{\agroup{1}}}{}=\FEval{\varfix{0},\varfix{1}}{\ips(\agroup{0})}$,
        \end{itemize}
        since all depend on the continuity of $\FEval{}{}$. This is also visually depicted in Figure~\ref{fig_teo}.\\
        This means there is also a cherry-picking model $\model_{\agroup{0}}\colon\agroup{0}\to\binaryset$ such that 
        \begin{equation*}
            \FEval{\model_{\agroup{1}}}{}
            =
            \FEval{\varfix{0},\varfix{1}}{\ips(\agroup{0})}
            = 
            \FEval{\model_{\agroup{0}}}{}
        \end{equation*}
        We are now ready to find an evaluation metric that forces cherry-picking. Consider an evaluation metric $\EEval{}{}$, independent of the distribution, with the following properties:
        \begin{itemize}
            \item $\EEval{}{}$ has a global maximum in $\model_{\agroup{1}}$ for $\agroup{0}$ 
            \item $\EEval{}{}$ has a global maximum in $\model_{\agroup{0}}$ for $\agroup{0}\cap 
            \{
                \model\colon \agroup{0}\to \binaryset\colon\FEval{\model}{}= \FEval{\model_{\agroup{1}}}{}
            \}$.
        \end{itemize}
        Notice that the second global maximum is taken on a specific level set of $\FEval{}{}$, which doesn't conflict with the first global maximum requirement.\\
        The existence of an evaluation metric with these properties is a consequence of Theorem~\ref{teo_freedom_of_loss} proven in the Appendix.\\
        Since $\agroup{0}$ and $\agroup{1}$ are a partition of $\featureset $, we can rewrite the distribution on $\featureset $ as such $\symmeasure_\featureset = (1-\varepsilon')\symmeasure_{\agroup{1}}+\varepsilon'\symmeasure_{\agroup{0}}$. 
        We can now prove that there exists constants $\varepsilon',c$ such that the following fairness problem
        \begin{equation*}
            \FairnessProblem{c\cdot}
        \end{equation*}
        has only cherry-picking solutions: if we take $\varepsilon'$ and $c$ indefinitely small we can notice that any solution $\optimalmodel $ that maximizes the problem must satisfy 
        \begin{equation*}
            \FEval{\optimalmodel\restricted{\agroup{0}}}{}
            =
            \FEval{\optimalmodel\restricted{\agroup{1}}}{}
            \quad\text{and} \quad
            \EEval{\optimalmodel}{}
            =
            \max_{\optimalmodel_\agroup{1}\colon \agroup{1}\to \binaryset}\EEval{\optimalmodel_\agroup{1}}{}
        \end{equation*}
        which means that $\optimalmodel \restricted{\agroup{0}}$ must cherry-pick, since by the properties of $\EEval{}{}$ we know that $\optimalmodel \restricted{\agroup{0}} = \model_{\agroup{0}}$. Since the solution to the problem is continuous in $\varepsilon'$ and $c$, and all the properties shown hold true for a suitable neighborhood of $\optimalmodel $, we can conclude that there exists also a $\varepsilon'$ and $c$ such that the problem has only cherry-picking solutions.
    \end{proof}
\end{theorem}
	There are several consequences of this result:

\begin{itemize} 
	\item This extends the results of \cite{baumann2022enforcing}, showing that cherry-picking might be necessary to satisfy fairness constraints defined on strict evaluation metrics. Predictive Parity, which uses Precision to evaluate fairness, falls into this category. 
	\item We know that the function $\FEval{}{}$ for Equal Opportunity and Equal Risk is an evaluation metric. However, since $\FEval{}{}$ is not strict, they do not satisfy the hypothesis of Theorem~\ref{teo_bad_fairness}, and thus avoid its consequences. 
	\item If a strict evaluation $\FEval{}{}$ guarantees non-cherry-picking solutions, then it means that $\FEval{}{}$ is constant for all Pareto-optimal models. This might seem like a strange condition, but it's actually quite intuitive: an evaluation metric that is constant on optimal models essentially only measures whether the model is Pareto-optimal or not. Given the connection shown in Theorem~\ref{teo_atomic_cherry_picking}, when $\FEval{}{}$ is used as a utility function, the fairness measure essentially penalizes any model that is not Pareto-optimal for a sensitive group. In other words, if there is aleatoric uncertainty, the fairness measure is directly measuring cherry-picking, and it is zero if and only if the model does not cherry-pick. 
	\item The last point allows us to rephrase the result in a more intuitive way: if $\featureset$ has aleatoric uncertainty and $\FEval{}{}$ is a strict evaluation metric, either we are actively avoiding cherry-picking, or there are no guarantees that a solution to the fairness problem does not cherry-pick. 
	\item The theorem is not exhaustive, as it only discusses the case of strict evaluation metrics. Additionally, the direction of the implication of the theorem is not reversible, since it is possible for a fairness measure to be constant on the Pareto-boundary and still have fairness problems with only cherry-picking solutions.
		
	The main reason is that the theorem only requires hypotheses on the classification problem $\featureset$ at hand. The continuity of $\FEval{}{}$ gives us information about the behavior of the fairness measure only locally for problems similar to $\featureset$. However, it is possible that for some sensitive groups $\{\agroup{0}, \agroup{1}\}$, the problem becomes very different, and $\FEval{}{}$ changes completely. This means that it is very difficult to completely characterize the fairness measures that might lead to cherry-picking without requiring additional assumptions.
	\end{itemize}
	Let's now revisit Example~\ref{example:cherry_picking}: requiring the same precision on both sensitive groups means that an optimal solution may exhibit cherry-picking, since precision is a strict evaluation metric. If we make the example a bit more numerical, we can use it to illustrate what can happen. This is similar to \cite{baumann2022enforcing}, which also provides some experiments on this specific case.  
	\medskip
	\begin{example}[Example \ref{example:cherry_picking} Revisited]\label{example_revisited}
		In the example, we were not allowed to change the precision of the model for the male group. In the theorem, this is equivalent to saying that the number of male candidates is vastly superior to the number of female candidates, so changing an optimal model for the male group would have a much larger negative impact than changing it for the female group.
		\begin{figure}[t!]
	\centering
	\includegraphics[width=.5\textwidth]{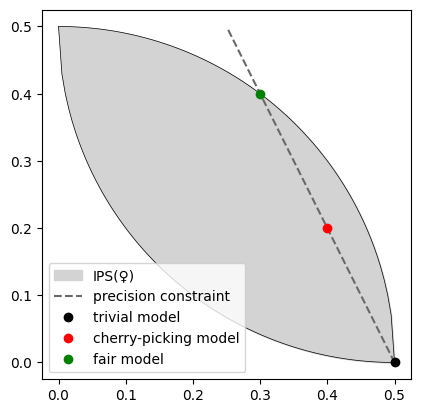}
	\caption
	{
		Visual aid for Example \ref{example_revisited}. The line is the constraint obtained by $k = 2/3$. 
        When $t>k$, the Immediate Utility $U_t(\model)$ is better for the cherry-picking model than the fair one. The trivial model is optimal, but other evaluation metrics might prefer the cherry-picking model. 
	}\label{fig_example_cherry}
\end{figure}
		So, we can set the precision of the model for the male group as $\prob(\labely =1\conditional \model=1, A=\agroup{1})= k$.

		We want the model to have exactly the same precision for the female group, which is a fairness constraint. In the theorem, this is equivalent to saying that the fairness measure is infinitely more important than the evaluation metric. We know that any model $\model_{\agroup{0}}\colon \agroup{0} \to \binaryset$ must satisfy the following linear equation:
		\begin{equation*}
			\varrate{1} = \frac{-k}{1-k}\left(\varrate{0}-\prob(\labely =0\conditional A=\agroup{0})\right)
		\end{equation*}
		where $\varrate{0} = \jointrate{0}{\model_{\agroup{0}}}{A=\agroup{0}}$ and $\varrate{1}= \jointrate{1}{\model_{\agroup{0}}}{A=\agroup{0}}$.

		The university staff now proposes to use Immediate Utility as an evaluation metric, which means maximizing the following objective:
		\begin{equation*}
			{U}_t(\model) = (1-t)\cdot\varrate{1}+t\cdot(\varrate{0}
			-\prob(\labely = 0\conditional A={\agroup{0}}))
		\end{equation*}
		We can now write a linear problem that perfectly represents the situation:
		\begin{equation*}
				\max_{(\varrate{0},\varrate{1})\in \ips(\agroup{0})}\quad \varrate{1}+\frac{t}{1-t}\cdot\varrate{0}\quad
				\text{s.t.}\quad \varrate{1} = \frac{-k}{1-k}\left(\varrate{0}-\prob(\labely =0\conditional A=\agroup{0})\right)
		\end{equation*}
		Which is the same as
		\begin{equation*}
			\max_{(\varrate{0},\varrate{1})\in \ips(\agroup{0})}\quad \left(\frac{t}{1-t}-\frac{k}{1-k}\right)\cdot\varrate{0}
		\end{equation*}
		In particular, if the university chooses $t > k$, the optimal solution is to maximize $\jointrate{0}{\model_{\agroup{0}}}{A = \agroup{0}}$ regardless of $\jointrate{1}{\model_{\agroup{0}}}{A = \agroup{0}}$, which means that cherry-picking solutions are preferable to the non-trivial and non-cherry-picking model, as illustrated in Figure \ref{fig_example_cherry}.

In this specific case, the trivial solution is optimal, meaning not selecting a single female candidate. The fact that the trivial solution is best is a consequence of the choice of Immediate Utility. Other evaluation metrics might lead to different results and force cherry-picking. Nonetheless, Immediate Utility demonstrates problematic behavior as predicted by Theorem~\ref{teo_bad_fairness}, and it is sufficient to illustrate what can go wrong.
	\end{example}
\section{Conclusions, Future Work and Limitations}
The results we have shown in this paper are quite formal and precise, so before continuing with the conclusions we would like to specify what they are \emph{not} claiming.

First of all, we are not claiming that all optimal solutions of every fairness problem must cherry-pick. This is formally false, since we have shown that for some common fairness measures there are always non-cherry-picking solutions. Also, for the other fairness measures, this statement doesn't hold true in general, and it very much depends on the specific details of the problem at hand. For instance, if in our fairness problem we use a constant evaluation metric, we shouldn't expect any cherry-picking.

Similarly, we are not claiming that for any sensitive partition of the population the cherry-picking phenomenon happens. We have shown that for some partitions, cherry-picking could be unavoidable if we simply optimize a model.

What we \emph{are} claiming is that for a number of fairness measures and fairness problems, there is no guarantee that cherry-picking can be avoided. The problem is that a priori it is often not clear if a fairness problem can guarantee at least a non-cherry-picking solution, which means that practitioners should be extremely careful before optimizing a fairness measure, since they could end up with cherry-picking solutions.

That said, we believe it would be interesting to investigate how often cherry-picking occurs in practice and if there are some fairness measures that are more prone to it. This is not an easy task in practice, as it would require almost perfect knowledge of the underlying distribution and is very dependent on the quality of the data. Some literature has been developed in that direction \cite{baumann2022enforcing, goethals2024beyond}, but the problem still remains an interesting topic for future research.

That being said, we do expect this phenomenon to arise naturally sufficiently often: the reasoning behind the proof of Theorem~\ref{teo_bad_fairness} requires finding a minority where the quality of models is strictly worse than the majority. This is a common situation in many real-world problems, where data collected from marginalized groups are often of lower quality due to bias and discrimination.

Moreover, once again, our results show how people with malicious intentions can exploit the fairness problem to appear fair on one hand, while unfairly treating some groups on the other hand. They could even justify their choices as being "theoretically optimal."

Furthermore, the cherry-picking problem probably extends to scores as well. If we manually set thresholds for each sensitive group to avoid cherry-picking, based on scores optimized for fairness, we could still inadvertently cherry-pick. A model could potentially unfairly distribute scores, ranking less-deserving individuals higher than more-deserving ones, all while adhering to the fairness constraint. Consequently, any decision based on these scores might seem non-cherry-picking solely because the true scores are obscured by the model.

In light of these results, what we advocate is to stop viewing fairness as something to optimize for. Instead, practitioners should prioritize building reliable classifiers capable of providing well-calibrated scores. This ensures that all the information contained in the data, even if biased, can be collected, and that the type and extent of bias in the distribution can be measured to the best of our abilities.

Then, fairness policies should be considered as a post-processing step, where the goal is either to correct the biases in the data to achieve the most equitable outcome possible, or implementing affirmative action policies to guarantee a more equitable world for future generations. This is transparent, easier to apply, and, most importantly, it doesn't hide any unfairness behind a veil of optimality.

\section*{Declarations}
\textbf{Acknowledgments:}
We thank the AXA joint research initiative for their support. We thank dr. Daphne Lenders for providing feedback and Example~\ref{example:cherry_picking} in Section~\ref{sec_understanding_cherry}. 

\begin{itemize}
\item Funding: First author is supported by the AXA joint research initiative (CS15893). 
\item Conflict of interest: No conflict of interest outside authors' institution.
\item Ethics approval: Not applicable
\item Consent to participate: Not applicable
\item Consent for publication: Not applicable
\item Availability of data and materials: Not applicable
\item Code availability: Not applicable
\item Authors' contributions: All authors contributed to the study conception and design. Formal proofs were originally discovered by the first author and double-checked by the second. First draft of the manuscript written by the first author while second author provided feedback on the manuscript. All authors read and approved the final manuscript.
\end{itemize}
\bibliography{biblio}
\pagebreak
\section*{Appendix}
\begin{definition}[Strictly Increasing]\label{def_first_quadrant}
  Let $U\subseteq\RR^n$ be an open set and $f\colon U\to \RR$ a function.
  We say that the function $f$ is strictly increasing at point $\vectorize{x}$ if for all $\vectorize{y}\in U$, it holds
  \begin{equation*}
    \vectorize{x} < \vectorize{y} \Rightarrow f(\vectorize{x})< f(\vectorize{y})
  \end{equation*} 
\end{definition}
\begin{theorem*}
	Let $f\colon [0,1]^n\to \RR$ be a $C^{1}$ function. If
	\begin{equation*}
	\forall \vectorize{x}\in [0,1]^n \quad \exists i\in[n]\text{ such that }
	{\partial_{x_i}\, f}(\vectorize{x})\geq 0\quad 
	\end{equation*}
	then
	\begin{equation*}
	\exists\, \vectorize{x}\in [0,1]^n\smallsetminus [0,1)^n \text{ such that }
	f(\vectorize{0})\leq f(\vectorize{x})
	\end{equation*}
	\begin{proof}
		We are going to prove this via induction on the dimension $n$.\\
		\textbf{Case} $n=1$\\
		In this case the theorem simply states that the function $f\colon [0,1]\to \RR$ has non-negative derivative and so it's non-decreasing. Then clearly $f(1)\geq f(0)$.\\
		\textbf{Case} $n-1\to n$\\
		Let's now consider the set $Q$ of points of $[0,1)^n$ where $f$ is strictly decreasing, formally
		\begin{equation*}
			Q:=\big\{\vectorize{x}\in [0,1)^n\colon f(\vectorize{x})>f(\vectorize{y})\ \forall \vectorize{y}\in [0,1]^n \text{ such that }\vectorize{y}>\vectorize{x} \big\}
		\end{equation*}
		$Q$ exhibits the following curious property that we can exploit:
		\begin{claim*}
			\begin{equation*}
			\vectorize{q}\in Q\ \Rightarrow\ \nabla f(\vectorize{q})= \vectorize{0} 
			\end{equation*}
			\begin{proof}[Proof of the claim]
				Via contradiction, let's assume the claim to be false. Then, there exists $i\in[n]$ such that $\partial_{x_i}\,f(\vectorize{q})\neq 0$. We can safely assume that $i=1$.\\
				If $\partial_{x_1}\,f(\vectorize{q})> 0$ we can find a $\varepsilon>0$ so that the point $\overline{\vectorize{q}} :=(q_1+\varepsilon,q_2,\dots,q_n )$ is such that $f(\overline{\vectorize{q}})\geq f(\vectorize{q})$. Since $\vectorize{q}\in [0,1)^n$ we can choose an $\varepsilon$ small enough such that $\overline{\vectorize{q}}\in [0,1)$, but then by definition we have $\vectorize{q}\not\in Q$ which is a contradiction.\\
				Otherwise if $\partial_{x_1}\,f(\vectorize{q})< 0$, since all partial derivatives are continuous, we can find an $\varepsilon>0$ such that 
				\begin{equation*}
				\vectorize{x}\in [q_1,q_1+\varepsilon]\times \dots\times [q_n,q_n+\varepsilon]\Rightarrow \partial_{x_1}\,f(\vectorize{x})< 0
				\end{equation*}
				Consider the function
				\begin{align*}
					g\colon [0,1]&^{n-1}\to \RR\\
					(x_2,\dots,&x_n)\mapsto 
					f(q_1,q_2+\varepsilon x_2,\dots,q_n+\varepsilon x_n)
				\end{align*} 
				Now, $g$ must satisfy all the hypothesis of the proposition, in particular there always is $i\in \{2,\dots, n\}$ such that $\partial_{x_i}\,g(\vectorize{x})\geq 0$ in $[0,1]^{n-1}$, otherwise we would have a point where $\partial_{x_i}\,f<0$ for all $i\in [n]$. By induction we can find a point $\vectorize{q}':=(q'_2,\dots, q'_n)\in [0,1]^{n-1}\smallsetminus [0,1)^{n-1}$ such that $g(\vectorize{0})\leq g(\vectorize{q}')$.\\
				In particular if we now consider the point $\overline{ \vectorize{q}}:=(q_1,q_2+\varepsilon q' _2,\dots,q_n+\varepsilon q' _n)$, it is such that $f(\overline{\vectorize{q}})\geq f(\vectorize{q})$. But this is a contradiction, since it would imply ${\vectorize{q}}\not\in Q$.
			\end{proof}
		\end{claim*}
		The claim says that $Q$ is a subset of the set of stationary points of $f$. This is important because the set of stationary points $\{\nabla f = \vectorize{0}\}$ in $[0,1]^n$ is a compact set since $\nabla f$ is a continuous function and $\{\vectorize{0}\}$ is a closed set. \\
		Because of this we can build a finite covering $\mathcal{F}_{\delta}:=\big\{H^{(t)}\big\}_{t\in I}$ of $\{\nabla f = \vectorize{0}\}$, depending on a value $\delta>0$, made of hypercubes $H^{(t)}:=H(\vectorize{\alpha}^{(t)}, r^{(t)})$ such that $r^{(t)}<\delta$ for all $H^{(t)}\in \mathcal{F}_\delta$, where an hypercube is 
		\begin{equation*}
		H(\vectorize{\alpha}, r):=\{\vectorize{x}\in \RR^n \colon \left\Vert \vectorize{x}-\vectorize{\alpha} \right\Vert_{\infty}<r\}=
		\{\vectorize{x}\in \RR^n \colon \left\vert {x}_i-\alpha_i \right\vert<r\quad \forall i\in[n]\}
		\end{equation*}
		The idea now is that for all $\varepsilon>0$ we will find a sequence $(\vectorize{x}^{(k)})_{k\in \NN}$ in $[0,1]^n$ starting at $\vectorize{x}^{(0)}:=0$ and converging to a point $\vectorize{x}'\in [0,1]^n\smallsetminus [0,1)^n$ such that $f(\vectorize{0})-f(\vectorize{x}')<\varepsilon$. Therefore, since $\varepsilon$ is arbitrary and $[0,1]^n\smallsetminus [0,1)^n$ is closed, this also implies that there exists $\overline{\vectorize{x}}\in [0,1]^n\smallsetminus [0,1)^n$ such that $f(\overline{\vectorize{x}})\geq f(\vectorize{0})$.\\ 
		We build our sequence using the following rules (and the axiom of choice): for every element $\vectorize{x}^{(k-1)}$ there are three possible situations, namely:
		\begin{itemize}
			\item $\vectorize{x}^{(k-1)}\in [0,1]^n\smallsetminus [0,1)^n$,
			\item $\vectorize{x}^{(k-1)}\in [0,1)^n\smallsetminus Q$,
			\item $\vectorize{x}^{(k-1)}\in Q$.
		\end{itemize}
		If $\vectorize{x}^{(k-1)}\in [0,1]^n\smallsetminus [0,1)^n$ we have already arrived to the set we want to converge to, so we simply pick $\vectorize{x}^{(k)}:=\vectorize{x}^{(k-1)}$.\\
		If $\vectorize{x}^{(k-1)}\in [0,1)^n\smallsetminus Q$ we know that the set 
		\begin{equation*}
		\big\{\vectorize{y}\in [x_1^{(k-1)},1]\times\dots\times [x_n^{(k-1)},1]\colon \vectorize{y}\neq \vectorize{x}^{(k-1)}\text{ and } f(\vectorize{x}^{(k-1)})\leq f(\vectorize{y})\big\}
		\end{equation*}
		is non-empty, so we can pick any element from that set as $\vectorize{x}^{(k)}$.\\
		If $\vectorize{x}^{(k-1)}\in Q$ then $\vectorize{x}^{(k-1)}\in H$ for some $H\in \mathcal{F}_\delta$. Since $H$ is a $n$-dimensional open interval as is $(0,1)^n$ there exist $\vectorize{a},\vectorize{b}\in \RR^n$ such that 
		\begin{equation*}
		H\cap (0,1)^n = (a_1,b_1)\times\dots\times (a_n,b_n)
		\end{equation*}
		In this case our choice for the next point in the sequence will be $\vectorize{x}^{(k)}:= \vectorize{b}$. If there are multiple elements of $\mathcal{F}_\delta$ we can choose from, any of them will work.
		Notice that any sequence $(\vectorize{x}^{(k)})_{k\in\NN}$ built in such a way is non-decreasing in any direction, that is 
		\begin{equation*}
		x^{(k)}_i\leq x^{(k+1)}_i \quad\forall i\in [n], k\in \NN
		\end{equation*}
		Also, only a finite amount of elements lie in $Q$.
		We can assume that the sequence $(\vectorize{x}^{(k)})_{k\in\NN}$ converges to the set $[0,1]^n\smallsetminus [0,1)^n$: suppose it doesn't, then, since the sequence is non-decreasing and bounded by $[0,1]^n$, it must converge to a point $\vectorize{p}\in [0,1)^n$. Now, since $\mathcal{F}_\delta$ is a finite set, we can be certain that in a neighborhood of $\vectorize{p}$ the points of the sequence are all outside $Q$, so there exists $m\in \NN$ such that
		\begin{equation*}
		f(\vectorize{x}^{(m)})\leq f(\vectorize{x}^{(m+1)})\leq\dots\leq f(\vectorize{x}^{(m+i)})\leq\dots\quad \forall i\in\NN 
		\end{equation*}
		Since $f$ is continuous we also have $f(\vectorize{x}^{(m)})\leq f(\vectorize{p})$. We can then redefine the sequence removing all the points after $\vectorize{x}^{(m)}$ and redefining $\vectorize{x}^{(m+1)}:= \vectorize{p}$, notice that this choice of $\vectorize{x}^{(m+1)}$ still follows the same rules previously stated. Continuing this new sequence we get a sequence that converges to a point strictly closer to $[0,1]^n\smallsetminus [0,1)^n$; it follows that the supremum of the possible point reachable in such a way is in $[0,1]^n\smallsetminus [0,1)^n$.
		So now we have a sequence $(\vectorize{x}^{(k)})_{k\in\NN}$ converging to $\vectorize{x}'\in [0,1]^n\smallsetminus [0,1)^n$. Also, by the same reasoning, the sequence can be chosen such that it is finite, so $(\vectorize{x}^{(k)})_{k\leq m}$ for some $m\in\NN$ and $\vectorize{x}^{(k)}\in [0,1)^n$ if and only if $k<m$. The only thing that remains to be proven is that we can choose this sequence in a way such that $f(\vectorize{0})-f(\vectorize{x}')<\varepsilon$.\\
		We have
		\begin{equation*}
		f(\vectorize{0})-f(\vectorize{x}') = f(\vectorize{x}^{(0)})-f(\vectorize{x}^{(m)})=\sum_{k\in [m]}f(\vectorize{x}^{(k-1)})-f(\vectorize{x}^{(k)})
		\end{equation*}
		We can split the sum according to the rules we have followed to build the sequence, since only $\vectorize{x}^{(m)}\in [0,1]^n\smallsetminus [0,1)^n$ one of the rules does not appear:
		\begin{equation*}
		\sum_{k\in [m]}f(\vectorize{x}^{(k-1)})-f(\vectorize{x}^{(k)})= \sum_{\substack {k\in [m]\\ \vectorize{x}^{(k-1)}\not\in Q}}f(\vectorize{x}^{(k-1)})-f(\vectorize{x}^{(k)}) + \sum_{\substack {k\in [m]\\ \vectorize{x}^{(k-1)}\in Q}}f(\vectorize{x}^{(k-1)})-f(\vectorize{x}^{(k)})
		\end{equation*}
		Since $f(\vectorize{x}^{(k-1)})-f(\vectorize{x}^{(k)})\leq 0$ if $\vectorize{x}^{(k-1)}\not\in Q$, the first term of the right-hand-side of the equation is always non-negative; it follows
		\begin{equation*}
		f(\vectorize{0})-f(\vectorize{x}')\leq \sum_{\substack {k\in [m]\\ \vectorize{x}^{(k-1)}\in Q}}f(\vectorize{x}^{(k-1)})-f(\vectorize{x}^{(k)})
		\end{equation*}
		So if we find a suitable bound for the sequence of points in $Q$ we can prove the theorem.\\
		First we notice that
		\begin{equation*}
		\sum_{k\in [m]} \left\Vert \vectorize{x}^{(k-1)}-\vectorize{x}^{(k)}\right\Vert_\infty\leq n
		\end{equation*}
		because
		\begin{equation*}
		\sum_{k\in [m]} \left\Vert \vectorize{x}^{(k-1)}-\vectorize{x}^{(k)}\right\Vert_\infty\leq \sum_{k\in [m]}\sum_{i\in [n]} \left\vert {x}^{(k-1)}_i-x^{(k)}_i\right\vert \leq \sum_{i\in [n]}\left\vert {x}^{(0)}_i-x^{(m)}_i\right\vert \leq n
		\end{equation*}
		so obviously the same bound holds for points in $Q$
		\begin{equation*}
		\sum_{\substack {k\in [m]\\ \vectorize{x}^{(k-1)}\in Q}} \left\Vert \vectorize{x}^{(k-1)}-\vectorize{x}^{(k)}\right\Vert_\infty\leq n
		\end{equation*}
		Now, for all $i\in [n]$ we know that $\partial_{x_i}\,f$ is a continuous function. But since $[0,1]^n$ is compact, $\partial_{x_i}\, f$ is also uniformly continuous. 
		As a consequence there exists $\delta>0$ such that for all $\vectorize{x},\vectorize{y}\in [0,1]^n$ 
		\begin{equation*}
		\left\Vert \vectorize{x}- \vectorize{y} \right\Vert_\infty<\delta \ \Rightarrow\ \left\Vert \nabla f(\vectorize{x})- \nabla f(\vectorize{y}) \right\Vert_\infty< \varepsilon/n^{1.5} 
		\end{equation*}
		If we use such a $\delta$ to define our family $\mathcal{F}_\delta$ then we have the following property: if $\vectorize{x}\in H\in \mathcal{F}_\delta$ then $\left\Vert \nabla f(\vectorize{x})\right\Vert_\infty< \varepsilon/n^{1.5}$.
		This is because by the definition of $\mathcal{F}_\delta$ there exists $\vectorize{y}\in \{\nabla f = \vectorize{0}\}$ such that
		\begin{equation*}
		\left\Vert \vectorize{x}- \vectorize{y} \right\Vert_\infty<\delta \ \Rightarrow\ \left\Vert \nabla f(\vectorize{x})\right\Vert_\infty = \left\Vert \nabla f(\vectorize{x})- \nabla f(\vectorize{y}) \right\Vert_\infty< \varepsilon/n^{1.5}
		\end{equation*}
		That means that if we consider any $H\in \mathcal{F}_\delta$ we have that the gradient is everywhere bounded in $H$. Since $H$ is also convex, the function $f$ is Lipschitz in $\overline H$ with constant $\varepsilon/n$ according to the $\infty$-norm. This can be seen as follows: if $\vectorize{a},\vectorize{b}\in \overline H$ then if we consider the curve 
		$\gamma(t):= t\vectorize{b}+(1-t)\vectorize{a}$
		\begin{equation*}
		\left\vert f(\vectorize{a})-f(\vectorize{b})\right\vert=
		\left\vert \int_0^1 \nabla f(\gamma(t))\cdot \gamma' (t) \diff t\right\vert \leq 
		\int_0^1\left\vert \nabla f(\gamma(t))\cdot \gamma' (t) \right\vert \diff t
		\end{equation*}
		by Cauchy-Schwartz $\left\vert \nabla f(\gamma(t))\cdot \gamma' (t) \right\vert \leq\left\Vert \nabla f(\gamma(t))\right\Vert\cdot\left\Vert \gamma' (t) \right\Vert$ so
		\begin{equation*}
		\int_0^1\left\vert \nabla f(\gamma(t))\cdot \gamma' (t) \right\vert \diff t<
		\varepsilon/n^{1.5}\cdot\int_0^1 \left\Vert\gamma' (t)\right\Vert \diff t=\varepsilon/n^{1.5} \left\Vert a-b\right\Vert\leq\varepsilon/n \left\Vert a-b\right\Vert_\infty
		\end{equation*}
		since $\left\Vert \vectorize{a}-\vectorize{b}\right\Vert\leq \sqrt{n}\left\Vert \vectorize{a}-\vectorize{b}\right\Vert_\infty$.
		Putting everything together, since if $\vectorize{x}^{(k-1)}\in Q$ then there exists $H\in \mathcal{F}_\delta$ such that $\vectorize{x}^{(k-1)},\vectorize{x}^{(k)}\in \overline H$, we have
		\begin{equation*}
		f(\vectorize{0})-f(\vectorize{x}')\leq \sum_{\substack {k\in [m]\\ \vectorize{x}^{(k-1)}\in Q}}f(\vectorize{x}^{(k-1)})-f(\vectorize{x}^{(k)})< \sum_{\substack {k\in [m]\\ \vectorize{x}^{(k-1)}\in Q}} \varepsilon/n \left\Vert \vectorize{x}^{(k-1)}-\vectorize{x}^{(k)}\right\Vert_{\infty}\leq \varepsilon
		\end{equation*}
		proving the theorem.
	\end{proof} 
\end{theorem*}
\begin{theorem}\label{teo_analisi}
  Let $U\subseteq \RR^n$ be an open set and $f\colon U\to \RR$ a $C^1$ function.\\
  If $f$ is strictly increasing at a point $\vectorize{x}$, then there exists a point $\vectorize{y}\geq \vectorize{x}$ such that 
  \begin{equation*}
  {\partial_{x_i} f} (\vectorize{y})>0\quad \forall\ i\in[n]
  \end{equation*}
\begin{proof}
Since $U$ is open we can find an $\varepsilon>0$ such that 
$
[x_1,x_1+\varepsilon]\times\dots\times[x_n,x_n+\varepsilon]\subseteq U
$.
Consider the function
\begin{align*}
  g\colon [0&,1]^{n}\to \RR\\
  \vectorize{t}&\mapsto 
  -f(\vectorize{x}+\varepsilon \vectorize{t})
\end{align*} 
If, by contradiction, for all $\vectorize{y}\neq \vectorize{x}$ such that $i\geq x_i$ for all $i\in [n]$, we have $\partial_{x_j} f(\vectorize{y})\leq 0$ for some $j\in[n]$, then $g$ satisfies the hypothesis of the previous proposition. So there exists $\vectorize{q}\in [0,1]^n\smallsetminus [0,1)^n$ such that $g(\vectorize{0})\leq g(\vectorize{q})$, but then $f(\vectorize{x}+\varepsilon \vectorize{q} )\leq f(\vectorize{x})$. So $f$ is not strictly increasing at $\vectorize{x}$, which is a contradiction.
\end{proof}
\end{theorem}
\begin{theorem}\label{teo_almost_egregium}
	Let $f\colon (0,1)^2\to \RR$ be a $C^1$ function. It holds
	\begin{equation*}
		\text{If}\quad\partial_x f \cdot \partial_y f = 0 \quad\text{then}\quad \partial_x f =0 \quad\text{or}\quad \partial_y f = 0
	\end{equation*}
	\begin{proof}
		WLOG suppose $f$ is not constant, then there exists a point $(x_0,y_0)$ where the two partial derivatives are not both zero at the same time. So let's assume WLOG $\partial_x f (x_0,y_0)>0$ and $\partial_y f (x_0,y_0)=0$ (the case with $<$ is equivalent). Since $f$ is a $C^1$ function, by the continuity of $\partial_x f$, we know that there exists an open square neighborhood $I^2$ of $(x_0,y_0)$ where $\partial_x f>0$, which means by hypothesis that also $\partial_y f =0$ in that neighborhood.	
		We can show that $\partial_x f=\partial_x f(x_0,y_0)$ in $\{x_0\}\times (0,1)$: consider the value
		\begin{equation*}
			\overline{y}:=\inf\{y>y_0 \colon \partial_y f(x_0,y)\neq 0\}=\inf\{y>y_0 \colon \partial_y f(x_0,y)\neq 0\text{ and } \partial_x f(x_0,y)=0\}
		\end{equation*}
		clearly $\partial_x f (x_0,\overline{y}) = 0$, but also $\partial_x f (x_0, y_0)=\partial_x f (x_0, \overline{y})$ since $\forall y\in [y_0,\overline{y}]$
		\begin{equation*}
			\partial_y \left(\partial_x f (x_0, y)\right) = \partial_x \left(\partial_y f (x_0, y)\right) = \partial_x( 0) = 0
		\end{equation*}
		Notice that the partial derivatives commute since one is constant as proven in \cite{rudin1976principles}.
		But $\partial_x f (x_0, y_0)>0$, so $\overline{y}$ cannot exist. A similar reasoning using the supremum proves that $\partial_y f = 0$ in $I\times (0,1)$ where it holds $\partial_x\partial_y f = \partial_y \partial_x f = 0$. So, if $\partial_x f(x_0,y_0)\neq 0$, then $\partial_x f= \partial_x f(x_0,y_0)$ in $\{x_0\}\times (0,1)$. Now, if there exists also a point $(x_1,y_1)$ where $\partial_y f(x_1,y_1)\neq 0$, we have that $(x_0,y_1)$ has both partial derivatives different from zero, which is against the hypothesis. So $\partial_y f = 0$ in $(0,1)^2$.
	\end{proof}
\end{theorem}
\begin{corollary}\label{cor_egregium_connected}
	Let $f\colon U^2\to \RR$ be a $C^1$ function on a connected open set $U$. If $f$ is never locally constant, then it holds
	\begin{equation*}
		\text{If}\quad\partial_x f \cdot \partial_y f = 0 \quad\text{then}\quad \partial_x f =0 \quad\text{or}\quad \partial_y f = 0
	\end{equation*}
	\begin{proof}
		We can cover $U$ with square open sets. For every square Theorem~\ref{teo_almost_egregium} holds, which means that every set has either $\partial_x f = 0$ or $\partial_y f = 0$. Let $U_0$ be the union of the sets where $\partial_x f = 0$ and similarly $U_1$ with $\partial_y f = 0$. The intersection $U_0\cap U_1$ must be empty since $f$ is never locally constant; but since $U$ is connected it means $U_0=\emptyset$ or $U_1=\emptyset$. 
	\end{proof}
\end{corollary}
\begin{theorem}\label{teo_freedom_of_loss}
  Let $F(x,y)\colon [0,1]^2\to \RR$ be a monotonically increasing function in $\individual$ and $y$. Consider a $C^1$ function $G(x,y)\colon [0,1]^2\to \RR$ such that for any point $(x,y)$ satisfying $F(x,y)=0$ it holds
  \begin{align*}
   \partial_x F(x,y) =0 \ \Rightarrow\ \partial_x G(x,y) \geq 0 \text{ in a neighborhood of }(x,y)\\
   \partial_y F(x,y) =0 \ \Rightarrow\ \partial_y G(x,y) \geq 0 \text{ in a neighborhood of }(x,y)
  \end{align*}
   then there exists a function $\widehat{G}(x,y)$ monotonically increasing such that $\widehat{G}(x,y)=G(x,y)$ on the set $F(x,y)=0$. 
  \begin{proof}
   Consider the function $\widehat{G}(x,y) := G(x,y)+\alpha\cdot F(x,y)$. Clearly $\widehat{G}(x,y)$ has the same value of $G(x,y)$ on the set $F(x,y)=0$. We can now choose $\alpha$ such that $\widehat{G}(x,y)$ is monotonically increasing. We can do this by imposing that the gradient of $\widehat{G}(x,y)$ is always positive. This means
   \begin{equation*}
    \nabla \widehat{G}(x,y) = \nabla G(x,y) + \alpha \nabla F(x,y) \geq 0
   \end{equation*}
   and so
   \begin{equation*}
    \alpha \geq \max_{F(x,y)=0}\max\left(-\frac{\partial_x G(x,y)}{\partial_x F(x,y)}, -\frac{\partial_y G(x,y)}{\partial_y F(x,y)}\right)
   \end{equation*}
   A positive $\alpha$ exists since the hypothesis and the compactness of $[0,1]^2$ imply that the right-hand side is always bounded from above.
  \end{proof}
 \end{theorem}
\begin{lemma}\label{lemma_splitting_ips}
	Consider a binary classification problem on $\featureset$. For all $\varepsilon>0$ there exists a partition $\{\agroup{0}, \agroup{1}\}$ of $\featureset $ having measure $\symmeasure_\featureset(\agroup{0}) = 1/2= \symmeasure_\featureset(\agroup{1})$ such that
	\begin{itemize}
		\item $\ips({\agroup{0}})\subseteq \ips({\agroup{1}})$
		\item $\partial\ips({\agroup{0}})\cap \partial\ips({\agroup{1}})\subseteq  \partial \big([0,\prob(\labely = 0)]\times [0,\prob(\labely = 1)]\big)$ 
		\item $\lambda(\ips(\agroup{1})\triangle\ips(\agroup{0}))<\varepsilon	$
	\end{itemize}
	where $\lambda$ is the Lebesgue measure on $[0,1]^2$.
	\begin{proof}
		Let's first forget about the $\varepsilon$ part and prove that we can find a partition $\{\agroup{0}, \agroup{1}\}$ that satisfies the other conditions.\\
		As mentioned in Section~\ref{subsec:the one cake} we can assume $\featureset =[0,1]=\Delta^2$ and $\prob(\labely =1\conditional\individual)= x$ by using the push-forward measure $\symmeasure_\Delta := \symmeasure_\featureset\circ \prob(\labely \conditional -)^{-1}$ (which we will assume being absolutely continuous with respect to the Lebesgue measure, but the result holds in general). 
		Let $a_1 := \sup \{ a\in[0,1]\colon \symmeasure_\Delta([a,1])=1/2\}$ and consider the function $b_a :=b(a)\colon [0,a_1]\to [0,1]$ defined as
		\begin{equation*}
			b(a)= \inf \{ {x}\in[0,1]\colon \symmeasure_\Delta([a,x])=1/2\}
		\end{equation*}
	Then the value $\prob(\labely =1\conditional\individual\in[a, b_a])$ is a continuous non-decreasing function as $a$ varies, in particular, it's easy to see that 
	\begin{equation*}
		\prob(\labely =1\conditional\individual\in[0, b_0])< \prob(\labely =1)<\prob(\labely =1\conditional\individual\in[a_1, 1])
	\end{equation*}
	so there must be an interval $[a_{\agroup{0}},b_{\agroup{0}}]$ such that $\prob(\labely =1\conditional\individual\in [a_{\agroup{0}},b_{\agroup{0}}])=\prob(\labely =1)$.\\
	We define $\agroup{0}$ as this newly found interval $\agroup{0}:= [a_{\agroup{0}},b_{\agroup{0}}]$ and $\agroup{1}:= \featureset\smallsetminus \agroup{0}$.\\
	Clearly, $\symmeasure_\Delta(\agroup{0})=\symmeasure_\Delta(\agroup{1})=1/2$ and $\prob(\labely =1\conditional A=\agroup{0})=\prob(\labely =1)$. Moreover, since it's known that given two events $E$ and $E'$ it holds
	\begin{equation*}
		E\indep E' \ \Rightarrow\ (\featureset \smallsetminus E)\indep E'
	\end{equation*}
	we can conclude also that $\labely \indep A$, since they are both binary random variables.\\
	Take now a non-trivial Pareto-optimal model $\model_{\agroup{0}}$ on $\agroup{0}$, we need to exhibit a model $\model_{\agroup{1}}$ on $\agroup{1}$ such that 
	\begin{align*}
		\jointrate{1}{\model_{\agroup{0}}}{A=\agroup{0}}&<\jointrate{1}{\model_{\agroup{1}}}{A=\agroup{1}}\\
		\jointrate{0}{\model_{\agroup{0}}}{A=\agroup{0}}&<\jointrate{0}{\model_{\agroup{1}}}{A=\agroup{1}}
	\end{align*} 
	This suffices to show that $\partial\ips(\agroup{0})\cap \partial\ips(\agroup{1})=\{(0,\prob(\labely =1)),(\prob(\labely =0),0)\}$.\\
	Since $\model_{\agroup{0}}$ is optimal and non-trivial, by Corollary~\ref{cor_binary_weller}, we know that there exists a threshold $a_{\agroup{0}}<\theta_{\agroup{0}}<b_{\agroup{0}}$ such that
	\begin{align*}
		\frac{1}{2}\jointrate{0}{\model_{\agroup{0}}}{A=\agroup{0}} &= \int_{a_{\agroup{0}}}^{\theta_{\agroup{0}}} \prob(\labely =0\conditional\individual) \symmeasure_\Delta(\diff )>0
		\\
		\frac{1}{2}\jointrate{1}{\model_{\agroup{0}}}{A=\agroup{0}} &= \int_{\theta_{\agroup{0}}}^{b_{\agroup{0}}} \prob(\labely = 1\conditional\individual) \symmeasure_\Delta(\diff\individual)>0		
	\end{align*}
	Consider now a new threshold $\theta_{\agroup{1}}$ such that
	\begin{equation*}
		\symmeasure_\Delta([0,\min(a_{\agroup{0}},\theta_{\agroup{1}})])+\symmeasure_\Delta([b_{\agroup{0}},\max(b_{\agroup{0}},\theta_{\agroup{1}})])= \symmeasure_\Delta([a_{\agroup{0}}, \theta_{\agroup{0}}])
	\end{equation*}
	which must exist since the dimension of $\agroup{0}$ and $\agroup{1}$ is the same. 
	We can now define a model $\model_{\agroup{1}}(x):= \mathbbm 1_{[\theta_{\agroup{1}},1]}(\prob(\labely = 1\conditional\individual))$ on $\agroup{1}$. We have now two cases:
	\medskip
	\\
	\textbf{Case 1:} $\theta_{\agroup{1}}\leq a_{\agroup{0}}$.
	In this case we have $\symmeasure_\Delta([0,\theta_{\agroup{1}}]) = \symmeasure_\Delta([a_{\agroup{0}}, \theta_{\agroup{0}}])>0$ so
	\begin{align*}
		\frac{1}{2}\jointrate{0}{\model_{\agroup{1}}}{A=\agroup{1}} =
		&\int_{0}^{\theta_{\agroup{1}}} \prob(\labely =0\conditional\individual) \symmeasure_\Delta(\diff\individual) =
		\\
		&\int_{0}^{\theta_{\agroup{1}}} (1-x) \symmeasure_\Delta(\diff\individual)>
		\\
		&\int_{0}^{\theta_{\agroup{1}}} (1-a_{\agroup{0}}) \symmeasure_\Delta(\diff\individual) = 
		\\
		&\int_{a_{\agroup{0}}}^{\theta_{\agroup{0}}} (1-a_{\agroup{0}}) \symmeasure_\Delta(\diff\individual)>
		\\
		&\int_{a_{\agroup{0}}}^{\theta_{\agroup{0}}} (1-x) \symmeasure_\Delta(\diff\individual) =
		\\
		&\int_{a_{\agroup{0}}}^{\theta_{\agroup{0}}} \prob(\labely =0\conditional\individual)\symmeasure_\Delta(\diff\individual)=
		\frac{1}{2}\jointrate{0}{\model_{\agroup{0}}}{A=\agroup{0}} 
	\end{align*}
	and similarly
	\begin{align*}
		\frac{1}{2}\jointrate{1}{\model_{\agroup{1}}}{A=\agroup{1}} =
		&\int_{[\theta_{\agroup{1}},1]\smallsetminus \agroup{0}} \prob(\labely = 1\conditional\individual) \symmeasure_\Delta(\diff\individual) =
		\\
		\prob(\labely =1)- &\int_{0}^{\theta_{\agroup{1}}} \prob(\labely = 1\conditional\individual)\symmeasure_\Delta(\diff\individual) >
		\\
		\prob(\labely =1)- &\int_{a_{\agroup{0}}}^{\theta_{\agroup{0}}} \prob(\labely = 1\conditional\individual)\symmeasure_\Delta(\diff\individual) =
		\\
		&\int_{\theta_{\agroup{0}}}^{b_{\agroup{0}}} \prob(\labely = 1\conditional\individual) \symmeasure_\Delta(\diff\individual) =
		\frac{1}{2}\jointrate{1}{\model_{\agroup{0}}}{A=\agroup{0}}
	\end{align*}
	\textbf{Case 2:} $\theta_{\agroup{1}}> a_{\agroup{0}}$.	In this other case we have $\symmeasure_\Delta([\theta_{\agroup{1}},1]) = \symmeasure_\Delta([\theta_{\agroup{0}},b_{\agroup{0}}])>0$, and the same reasoning as before holds by swapping the labels.\\
	This proves that $\ips(\agroup{0})\subset \ips(\agroup{1})$ with only trivial intersection. 
	We now need to prove that we can arbitrarily limit how big the difference between the two is. Fix $\varepsilon>0$,
	and consider a finite collection of points $\{x_i\}_{i=0\dots k}\subseteq \featureset$, satisfying
	\begin{equation*}
		0=x_0<x_1<\dots<x_{k}=1
	\end{equation*}
	such that $0<\symmeasure_\Delta([x_i,x_{i+1}])<\varepsilon/2$ for all $i\in [k-1]$.\\
	On every $\featureset _i:=[x_i,x_{i+1}]$, by conditioning on the interval, we can apply the same reasoning as before and get a subpartition $\{\agroup{0}_i, \agroup{1}_i\}$ satisfying the previous conditions. We can then take the union of all the $\agroup{0}_i$ and $\agroup{1}_i$ and define $\agroup{0}:=\bigcup_i \agroup{0}_i$ and $\agroup{1}:=\bigcup_i \agroup{1}_i$. \\
	Let's prove that $\agroup{0}$ and $\agroup{1}$ work as intended: first we need to show that they respect the condition for $\partial\ips(\agroup{0})\cap \partial\ips(\agroup{1})$. This follows easily because it's satisfied on every $\featureset _i$ which are almost everywhere disjoint, so for every non-trivial model on $\agroup{0}$ we can find a Pareto superior model on $\agroup{1}$ by carefully selecting how it behaves on all the $\agroup{1}_i$. Moreover we still have $\symmeasure_\Delta(\agroup{0})=\symmeasure_\Delta(\agroup{1})$ and $\labely\indep A$.\\
	Now we need to show that the difference between the two $\ips$s is less than $\varepsilon$ and strictly positive. 
	If $\model_{\agroup{0}}$ is Pareto-optimal on ${\agroup{0}}$, then as before we have a threshold $\theta_{\agroup{0}}$ that characterises $\model_{\agroup{0}}$. 
	In particular there exists ${j}$ such that $\theta_{\agroup{0}}\in [x_{{j}},x_{{j}+1}]=\featureset_{{j}}$. We then know that 
	\begin{align*}
	\jointrate{0}{\model_{\agroup{0}_i}}{A=\agroup{0}_i}=\prob(\labely =0\conditional \featureset_i)\text{ 
	for all }i<{j}\\
	\jointrate{1}{\model_{\agroup{0}_i}}{A=\agroup{0}_i}=\prob(\labely =1\conditional \featureset_i)\text{ for all }i>j
	\end{align*}
	Let's take now any optimal model $\model_{\agroup{1}}$ on $\agroup{1}$ such that 
	\begin{equation*}
		\jointrate{0}{\model_{\agroup{1}}}{A=\agroup{1}} = \jointrate{0}{\model_{\agroup{0}}}{A=\agroup{0}}
	\end{equation*} 
	and again consider the threshold $\theta_{\agroup{1}}$ that characterises it. We will prove that $\theta_{\agroup{1}}\in [x_{{j}},x_{{j}+1}]$ as well. Suppose they are not in the same interval and $\theta_{\agroup{0}}<\theta_{\agroup{1}}$. Then we would have
	\begin{align*}
		&\int_{\theta_{\agroup{0}}}^{1}\prob(\labely = 1\conditional\individual)\cdot\chara_{\agroup{0}}(x) \symmeasure_\Delta(\diff\individual)
		\geq
		\\
		&\int_{x_{{j}+1}}^{1}\prob(\labely = 1\conditional\individual)\cdot\chara_{\agroup{0}}(x) \symmeasure_\Delta(\diff\individual)
		=
		\\
		\sum_{i={j}+1}^{k-1}&\int_{x_i}^{x_{i+1}}\prob(\labely = 1\conditional\individual)\cdot\chara_{\agroup{0}_i}(x) \symmeasure_\Delta(\diff\individual)
		=
		\\
		\sum_{i={j}+1}^{k-1}&\int_{x_i}^{x_{i+1}}\prob(\labely = 1\conditional\individual)\cdot\chara_{\agroup{1}_i}(x) \symmeasure_\Delta(\diff\individual)
		= 
		\\
		&\int_{x_{{j}+1}}^{1}\prob(\labely = 1\conditional\individual)\cdot\chara_{\agroup{1}}(x) \symmeasure_\Delta(\diff\individual)
		>
		\\
		&\int_{\theta_{\agroup{1}}}^{1}\prob(\labely = 1\conditional\individual)\cdot\chara_{\agroup{1}}(x) \symmeasure_\Delta(\diff\individual)
	\end{align*}
	which is a contradiction since it would imply 
	\begin{equation*}
	\jointrate{1}{\model_{\agroup{0}}}{A=\agroup{0}}>\jointrate{1}{\model_{\agroup{1}}}{A=\agroup{1}}
	\end{equation*} 
	The same reasoning holds if $\theta_{\agroup{0}}>\theta_{\agroup{1}}$ by considering the negative label.\\
	As a consequence $\theta_{\agroup{0}}$ and $\theta_{\agroup{1}}$ are in the same interval $\featureset _{{j}}$, then by splitting the integral as we just did, we can see that
	\begin{multline*}
		\frac{1}{2}\big(\jointrate{1}{\model_{\agroup{1}}}{A=\agroup{1}} - \jointrate{1}{\model_{\agroup{0}}}{A=\agroup{0}}\big) = \\
		=\int_{\theta_{\agroup{1}}}^{x_{j+1}} \prob(\labely = 1\conditional\individual)\cdot\chara_{{\agroup{1}}_j}(x) \symmeasure_\Delta(\diff\individual) - \int_{\theta_{\agroup{0}}}^{x_{j+1}} \prob(\labely = 1\conditional\individual)\cdot\chara_{\agroup{0}_j}(x) \symmeasure_\Delta(\diff\individual) 
	\end{multline*}
	we already know that this quantity is strictly positive, but we can also bound it since
	\begin{align*}
		&\int_{\theta_{\agroup{1}}}^{x_{j+1}} \prob(\labely = 1\conditional\individual)\cdot\chara_{{\agroup{1}}_j}(x) \symmeasure_\Delta(\diff\individual) - \int_{\theta_{\agroup{0}}}^{x_{j+1}} \prob(\labely = 1\conditional\individual)\cdot\chara_{\agroup{0}_j}(x) \symmeasure_\Delta(\diff\individual)\leq
		\\
		&\int_{x_{j}}^{x_{j+1}} \prob(\labely = 1\conditional\individual)\cdot\chara_{{\agroup{1}}_j}(x) \symmeasure_\Delta(\diff\individual)\leq \int_{x_{j}}^{x_{j+1}} \chara_{{\agroup{1}}_j}(x) \symmeasure_\Delta(\diff\individual)=
		\symmeasure_\Delta(\featureset _j)/2<\varepsilon/4
	\end{align*}
	which implies $\jointrate{1}{\model_{\agroup{1}}}{A=\agroup{1}} - \jointrate{1}{\model_{\agroup{0}}}{A=\agroup{0}}<\varepsilon/2$.\\
	We know that there are two monotonically decreasing functions $f_{\agroup{0}}, f_{\agroup{1}}\colon [0,1]\to [0,1]$ such that for all $t\in [0,1]$ there exist Pareto-optimal models $\model_{\agroup{0}},\model_{\agroup{1}}$ such that 
	\begin{align*}
	&\big(\jointrate{0}{\model_{\agroup{0}}}{A=\agroup{0}}, \jointrate{1}{\model_{\agroup{0}}}{A=\agroup{0}}\big)= (t, f_{\agroup{0}}(t))\\
	&\big(\jointrate{0}{\model_{\agroup{1}}}{A=\agroup{1}}, \jointrate{1}{\model_{\agroup{1}}}{A=\agroup{1}}\big)= (t, f_{\agroup{1}}(t))\end{align*}
	Hence, because of the symmetry of the $\ips$ and since $\ips(\agroup{0})\subset \ips(\agroup{1})$ we have
	\begin{equation*}
		\lambda(\ips(\agroup{1})\triangle\ips(\agroup{0}))=\lambda(\ips(\agroup{1})) - \lambda(\ips(\agroup{0})) = 2\int_0^1 f_{\agroup{1}}(t) - f_{\agroup{0}}(t) \diff t
	\end{equation*}
	but the difference between $f_{\agroup{1}}$ and $f_{\agroup{0}}$ is exactly the difference of the probability of a true positive between the two optimal models on $\agroup{1}$ and $\agroup{0}$ sharing the same probability of a true negative. So
	\begin{equation*}
		0<2\int_0^1 f_{\agroup{1}}(t) - f_{\agroup{0}}(t) \diff t <2\int_0^1 \frac{\varepsilon}{2} \diff t = \varepsilon
	\end{equation*}
	\end{proof}
\end{lemma}
\begin{corollary}\label{cor_any_communities}
	Consider a binary classification problem on $\featureset$. 
	For all $(\varepsilon, \varepsilon')\in (0,1)^2$ there exists a partition $\{\agroup{0}, \agroup{1}\}$ of $\featureset $ having measure $\symmeasure_\featureset(\agroup{1}) = \varepsilon'$ and $\symmeasure_\featureset(\agroup{0}) = 1-\varepsilon'$ such that
	\begin{itemize}
		\item $\ips({\agroup{0}})\subseteq \ips({\agroup{1}})$
		\item $\partial\ips({\agroup{0}})\cap \partial\ips({\agroup{1}})\subseteq  \partial \big([0,\prob(\labely = 0)]\times [0,\prob(\labely = 1)]\big)$ 
		\item $\lambda(\ips(\agroup{1})\triangle\ips(\agroup{0}))<\varepsilon	$
	\end{itemize}
	where $\lambda$ is the Lebesgue measure on $[0,1]^2$.
	\begin{proof}
		It easy to see that if $\varepsilon'$ is a power of $1/2$ then we can simply apply the previous lemma multiple times in a row. If $\varepsilon'$ is not a power of $1/2$ then we can simply consider its binary expansion and apply the previous lemma on each digit.
	\end{proof}
\end{corollary}
\begin{lemma}\label{lemma_unique_problem}
	Let $\Delta$ and $\Delta'$ two binary classification problems on $[0,1]$. Then it holds:
	\begin{equation*}
		\ips(\Delta)=\ips(\Delta')\iff\symmeasure_{\Delta}=\symmeasure_{\Delta'}
	\end{equation*}
	\begin{proof}
		We just need to prove that the set of optimal points in $\ips(\Delta)$ uniquely identifies the distribution $\symmeasure_\Delta$. We prove it only for the case of $\symmeasure_\Delta$ being absolutely continuous with respect to the Lebesgue measure, but the result holds in general.\\
		Let $\varphi(x)\colon [0,1]\to [0,1]$ be the convex function describing the set of optimal points in $\ips(\Delta)$, we want to reconstruct the cumulative density function 
		\begin{equation*}
			F(T) = \int_0^T f(t)\diff t
		\end{equation*}
		for some density $f(x)$. Moreover, by Weller's theorem, we know that the set of optimal points in $\ips(\Delta)$ is described by the curve $(X(T), Y(T))$ where
		\begin{equation*}
			X(T) = \int_0^{T} (1-t) f(t)\diff t \quad
			Y(T) = \int_{T}^1 t\cdot f(t)\diff t
		\end{equation*}
		First we prove that the non-singleton intervals $\left[\underline{T}_{x},\overline{T}_{x}\right]$ satisfying
		\begin{equation*}
			\left[\underline{T}_{x},\overline{T}_{x}\right] = X^{-1}(\{x\})
		\end{equation*}
		for some $x\in [0,1]$ are at most countable. This is because at the point $x$ the function $\varphi'(x)$ is discontinuous or undefined, we can remove the cases where either $\underline{T}_{x}=0$ or $\overline{T}_{x}=1$ since they don't affect the countability claim.
		For all $\varepsilon>0$, we know that in $(\underline{T}_{x}-\varepsilon, \underline{T}_{x})$ the density $f(x)$ is not almost everywhere zero, otherwise $[\underline{T}_{x},\overline{T}_{x}]$ wouldn't be maximal. Moreover, by the Lebesgue differentiation theorem, we know that almost everywhere we have
		\begin{equation*}
			X'(T) = (1-T)\cdot f(T) \quad
			Y'(T) = -T\cdot f(T)
		\end{equation*}
		which means that we can find a $\underline{T}_\varepsilon\in (\underline{T}_x-\varepsilon, \underline{T}_x)$ such that $f(\underline{T}_\varepsilon)>0$ and
		\begin{equation*}
			\varphi'(\underline{x}_\varepsilon) = 
			\frac{-\underline{T}_\varepsilon\cdot f(\underline{T}_\varepsilon)}{(1-\underline{T}_\varepsilon)\cdot f(\underline{T}_\varepsilon)}
			=
			\frac{-\underline{T}_\varepsilon}{1-\underline{T}_\varepsilon}
		\end{equation*}
		for some $\underline{x}_\varepsilon$.
		Similarly, we can find a $\overline{T}_\varepsilon\in (\overline{T}_x, \overline{T}_x+\varepsilon)$ such that $f(\overline{T}_\varepsilon)>0$ and
		\begin{equation*}
			\varphi'(\overline{x}_\varepsilon) 
			= \frac{-\overline{T}_\varepsilon\cdot f(\overline{T}_\varepsilon)}{(1-\overline{T}_\varepsilon)\cdot f(\overline{T}_\varepsilon)} 
			=
			\frac{-\overline{T}_\varepsilon}{1-\overline{T}_\varepsilon} 
		\end{equation*}
		But then
		\begin{equation*}
			\lim_{\varepsilon\to 0} \varphi'(\underline{x}_\varepsilon) 
			= 
			\frac{-\underline{T}_\varepsilon}{(1-\underline{T}_\varepsilon)}
			\neq
			\frac{-\overline{T}_\varepsilon}{(1-\overline{T}_\varepsilon)} 
			=
			\lim_{\varepsilon\to 0} \varphi'(\overline{x}_\varepsilon)
		\end{equation*}
		which shows that $\varphi'(x)$ is discontinuous at $x$. Since $\varphi(x)$ is convex, the set of discontinuous points for the derivative is at most countable, which prove the countability of the intervals.\\ 
		We now know that, if for a point $x$ the derivative $\varphi'(x)$ is continuous and defined, then there exists a unique $T_x$ such that $X(T_x)=x$. Moreover, in every neighborhood of $T_x$ the density $f(x)$ is not almost everywhere zero. Using a similar argument as before, we can find points $x_\varepsilon\to x$ such that $f(x_\varepsilon)>0$ and
		\begin{equation*}
			\varphi'(x)=\lim_{\varepsilon\to 0}\varphi'(x_\varepsilon) = 
			\lim_{\varepsilon\to 0}
			\frac{-T_{x_\varepsilon}}{1-T_{x_\varepsilon}}
		\end{equation*}
		so given $\varphi'(x)$ we can find $T_x$ by solving
		\begin{equation*}	
		T_x = \frac{\varphi'(x)}{\varphi'(x)-1}	
		\end{equation*}
		Moreover, given the point $(x,\varphi(x))$, we have
		\begin{equation*}
			F(T_x) = x-\varphi(x)+\prob(\labely =1)
		\end{equation*}
		since
		\begin{equation*}
			x-\varphi(x) 
			=
			\int_0^{T_x} (1-t)f(t)\diff t - \int_0^1 t\cdot f(t)\diff t + \int_0^{T_x} t\cdot f(t)\diff t
			=
			F(T_x) - \prob(\labely =1)
		\end{equation*}
		So given a point $(x,\varphi(x))$, if $\varphi'(x)$ is continuous in $x$, we know 
		\begin{equation*}
			F\left(\frac{\varphi'(x)}{\varphi'(x)-1}\right) = x-\varphi(x)+\prob(\labely =1)
		\end{equation*}
		which defines $F$ uniquely.
	\end{proof}
\end{lemma}
\begin{lemma*}[\ref{lemma_eloss_exists}]\Paste{lemma_eloss_exists}
	\begin{delayedproof}{lemma_eloss_exists}
		Notice that the second point says that the output of $\symratedependent$ for models defined on the same decision problem only depend on the diagonal of the confusion matrix.
		This mean we can limit ourselves to prove the statement for models $\model$ and $\model'$ only depending on the value of $\prob_\featureset(\labely = 1\conditional \individual)$ and $\prob_{\featureset'}(\labely = 1\conditional \individual)$. Which then means that we can suppose that $\model$ and $\model'$ are function of the simplex $\Delta^2$.\\
		If $\jointrate{0}{\model}{}=\jointrate{0}{\model'}{}$, $\jointrate{1}{\model}{}=\jointrate{1}{\model'}{}$, and $\Delta_{\featureset}=\Delta_{\featureset'}$ for the two push-forward measures, then the two models are equivalent on $\Delta^2$, which would imply $\REval{\model}{}=\REval{\model'}{}$.
		But we proved in Lemma~\ref{lemma_unique_problem} that $\ips(\featureset )=\ips(\featureset')$ if and only if $\Delta_{\featureset}=\Delta_{\featureset'}$. So we have the thesis.
	\end{delayedproof}
\end{lemma*}
\begin{theorem*}[\ref{teo_extra_dim}]
	\Copy{clip_teo_extra_dim}{
	Consider a classification problem on $\featureset $. There exists a cake-cutting instance on $\featureset \times [0,1)$ such that for any matrix $M$, it holds:
	\begin{equation*}
		M = M( \model)\text{ for a model } \model \iff M=\vectorize{\symmeasure}(\vectorize{\slicing}) \text{ for some slicing } \vectorize{\slicing}=(\slicing_1,\dots, \slicing_n) 
	\end{equation*}
	}
	\begin{delayedproof}{teo_extra_dim}
		First, let's formalize the cake-cutting instance $(\featureset \times [0,1), \Sigma, \vectorize{\symmeasure})$ that we need.
		We already know that the cake is $\featureset \times [0,1)$, so a natural choice for the $\sigma$-algebra $\Sigma$ is the product algebra $\Sigma_\featureset\otimes \mathcal B([0,1))$, where $\Sigma_\featureset$ is the original $\sigma$-algebra on $\featureset $ and $\mathcal B([0,1))$ is the Borel $\sigma$-algebra on $[0,1)$.\\
		Similarly, $\vectorize{\symmeasure} = (\symmeasure_1,\dots,\symmeasure_n)$ is defined as the vector of the product measures $\symmeasure_i$ such that
		\begin{equation*}
			\symmeasure_i\left(Q\times [a,b] \right) := \lambda([a,b])\cdot{\int_Q \prob(\labely = i\conditional \individual) \symmeasure_\featureset(\diff \individual) } = (b-a)\cdot \prob(Q, \labely = i)
		\end{equation*}
		The use of the Lebesgue measure $\lambda$ in the product guarantees that $\symmeasure_i$ is an atomless measure. We can now prove that such a cake-cutting instance satisfies the claim of the proposition.\\
		Let's start by proving that, given a probabilistic model $ \model\colon \featureset\to \labelset$, we can construct a slicing $\vectorize{\slicing}$ for the cake-cutting instance such that $\symmeasure_i(\slicing_j) = \prob(\model = j,\labely = i)$ for all $i,j\in [n]$. We define $\slicing_j$ as follows:
		\begin{equation*}
		\slicing_j = \left\{ (\individual,t) \in \featureset\times [0,1)\colon \sum_{k=1}^{j-1}
		\prob(\model(\individual)=k)\leq t < \sum_{k=1}^{j}\prob(\model(\individual)=k)\right\}
		\end{equation*}
		Since $\sum_{k=1}^{n}\prob(\model(\individual)=k)=1$, we have that the slicing $\vectorize{\slicing} = (\slicing_1,\dots, \slicing_n)$ forms a partition of $\featureset \times [0,1)$. 
		They are also measurable, since the sub-graphs of measurable functions are measurable sets in $\featureset \times [0,1)$. Therefore, every $\slicing_j$ is measurable, as it can be expressed as the difference of measurable sets. 
		Consequently,
		\begin{equation*}
		\symmeasure_i(\slicing_j) = \int_{\slicing_j} \prob(\labely = i\conditional \individual) \symmeasure_\featureset(\diff \individual)\otimes \diff t 
		\end{equation*}
		By Fubini-Tonelli's theorem, the integral can be rewritten as
		\begin{equation*}
			\int_{\featureset}\left( \int_{\sum_{k=1}^{j-1}\prob(\model(\individual)=k)}^{\sum_{k=1}^{j}\prob(\model(\individual)=k)} \diff t\right)\prob(\labely = i\conditional \individual) \symmeasure_\featureset(\diff \individual)
			= 
			\int_{\featureset} \prob(\model(\individual)=j)\prob(\labely = i\conditional \individual) \symmeasure_\featureset(\diff \individual)
		\end{equation*}
		since the inner integral is just 
		\begin{equation*}
		\sum_{k=1}^{j}\prob(\model(\individual)=k)-\sum_{k=1}^{j-1}\prob(\model(\individual)=k)=\prob(\model(\individual)=j)
		\end{equation*}
		Finally,
		\begin{equation*}
		\int_{\featureset} \prob(\model(\individual)=j)\prob(\labely = i\conditional \individual) \symmeasure_\featureset(\diff \individual)
		= 
		\prob(\model = j,\labely = i)
		\end{equation*}
		On the other hand, if we start with a slicing $\vectorize{\slicing} = (\slicing_1,\dots, \slicing_n)$ of $\featureset \times [0,1)$, we can consider the probabilistic decision function $ \model(\individual)$ with probabilities defined as follows:
		\begin{equation*}
			\prob(\model(\individual)=j) = \int_0^1 \mathbbm 1_{\slicing_j}(\individual,t)\diff t
		\end{equation*}
		where $\mathbbm 1_{\slicing_j}$ is the characteristic function of the set ${\slicing_j}$. Since ${\slicing_j}$ is measurable, its characteristic function is also measurable. Then, again by Fubini-Tonelli's theorem, we have that $\prob(\model(\individual)=j)$ is a measurable function in $\featureset $, and it holds
		\begin{align*}
			&\int_\featureset \prob(\model(\individual)=j)\prob(\labely = i\conditional \individual) \symmeasure_\featureset(\diff \individual) = \\
			&\int_\featureset \left( \int_0^1 \mathbbm 1_{\slicing_j}(\individual,t)\diff t \right)\prob(\labely = i\conditional \individual) \symmeasure_\featureset(\diff \individual) =
  		\symmeasure_i(\slicing_j)
		\end{align*}
		Finally, $\model(\individual)$ is sampled from a proper distribution because
		\begin{equation*}
		\sum_{j=1}^n \prob(\model(\individual)=j) = \sum_{j=1}^n \int_0^1 \mathbbm 1_{\slicing_j}(\individual,t)\diff t = \int_0^1\sum_{j=1}^n \mathbbm 1_{\slicing_j}(\individual,t)\diff t = \int_0^1 \diff t=1
		\end{equation*}
		since $\sum_{j=1}^n\mathbbm 1_{\slicing_j}(\individual,t)=1$, given that $\vectorize{\slicing}$ is a partition of $\featureset \times [0,1)$.
	\end{delayedproof}
\end{theorem*}
\begin{theorem*}[\ref{teo_IPS_open}]\Paste{teo_IPS_open}
	\begin{delayedproof}{teo_IPS_open}
		If $ \model\colon \featureset\to \labelset$ is already Pareto-optimal, then by definition, there is no $\varepsilon>0$ that works.\\
		If $ \model\colon \featureset\to \labelset$ is not Pareto-optimal, then by the definition of Pareto-optimality, there exists $ \model'\paretostrictmore \model$ such that $[M( \model)]_{i,i} < [M( \model')]_{i,i}$ for some $i\in [n]$. 
		Because of Theorem~\ref{teo_extra_dim}, we can then find slicings $\vectorize{\slicing}, \vectorize{\slicing}'$ of $\featureset \times [0,1)$ equivalent to $ \model, \model'$, respectively.
		Let $Q\subseteq \slicing'_i$ be a set such that $\symmeasure_i(Q)= \symmeasure_i(\slicing'_i)-\symmeasure_i(\slicing_i)$. Such a set must exist because $\symmeasure_i$ is an atomless measure.
		In particular, $\symmeasure_j(Q)>0$ for all $j\in[n]$, since almost everywhere $\prob(\labely = j\conditional \individual)>0$. Because of this, we can construct slicings $\vectorize{\slicing}^{(j)}=(\slicing_1^{(j)},\dots,\slicing_n^{(j)})$ for all $j\in [n]$ defined as:
		\begin{equation*}
		\slicing_k^{(j)}:=
		\begin{cases}
			\slicing'_k\smallsetminus Q &\text{if }k\neq j\\
			\slicing'_k\cup Q &\text{if }k= j
		\end{cases}
		\end{equation*}
		Notice that $\vectorize{\slicing}\paretostrictless\vectorize{\slicing}^{(j)}$ and $\symmeasure_j(\slicing_j)< \symmeasure_j(\slicing_j^{(j)})$ for all $j\in [n]$. Due to Theorem~\ref{teo_Dvoretsky}, this means that there exists $\varepsilon>0$ such that
		\begin{equation*}
		\left\{\diag \vectorize{\symmeasure}(\vectorize{\slicing})\right\} +[0,\varepsilon]^n\subseteq \ips(\featureset \times [0,1))
		\end{equation*}
		Hence, once again, we have the thesis by applying Theorem~\ref{teo_extra_dim}.
	\end{delayedproof}
\end{theorem*}
\begin{corollary*}[\ref{cor_binary_weller}]\Paste{cor_binary_weller}
	\begin{delayedproof}{cor_binary_weller} 
	A model $\model$ is Pareto-optimal if and only if there exists a sequence $\vectorize{{w}}^{(k)}=({w}_0^{(k)},{w}_1^{(k)})_{k\in\NN}$ associated with the model. 
	Thus, we need to show that the models described by the theorem are exactly those associated with such a sequence.\\
	Since $\prob(\labely =0\conditional \individual)+\prob(\labely = 1\conditional \individual)=1$ and similarly ${w}^{(k)}_0 + {w}_1^{(k)}=1$, for any sequence in $\Delta^2\cap (0,1)^2$ we have
	\begin{equation*}
		\frac{\prob(\labely = 1\conditional \individual)}{\prob(\labely =0\conditional \individual)}\geq \frac{{w}_1^{(k)}}{{w}_0^{(k)}} \iff \prob(\labely = 1\conditional \individual)\geq {w}_1^{(k)}\iff \prob(\labely =0\conditional \individual)\leq {w}_0^{(k)}
	\end{equation*}
	For the same reason, the limit $\lim_{k\to \infty} {{w}_0^{(k)}}/{{w}_1^{(k)}}$ converges in $[0,\infty]$ if and only if $\lim_{k\to \infty} {{w}_1^{(k)}}$ converges in $[0,1]$.\\ 
	Let's now consider a model $\model$ associated with a sequence $\vectorize{{w}}^{(k)}$.
	Now, Weller's theorem, even in the non-binary case, ties our hands for points where $\prob(\labely = i\conditional \individual)=0$ for some $i$. In that case a.e. $\prob(\model(\individual)=i)=0$; otherwise, $\prob(\model(\individual)=i)>0$ would imply that 
	\begin{equation*}
		0 = \frac{\prob(\labely = i\conditional \individual)}{\prob(\labely = j\conditional \individual)}\geq \frac{{w}_i^{(k)}}{{w}_j^{(k)}}>0
	\end{equation*}
	which, independently of the sequence $\vectorize{{w}}^{(k)}$, always results in a contradiction. 
	\\
	The other points instead depend on the sequence, so let's call $t:=\lim_{k\to \infty} {{w}_1^{(k)}}$. If $\prob(\labely = 1\conditional \individual)<t$, then for the same reasoning as above, $\prob(\model(\individual)=1)= 0$. Similarly, if $\prob(\labely = 1\conditional \individual)>t$, then $\prob(\model(\individual)=1)=1$.
	\\
	The only points where $\model(\individual)$ is not predetermined by Weller's theorem are exactly those where $\prob(\labely = 1\conditional \individual)=t$, and they can be freely assigned using a function $q(x)$. Hence, we have that
	\begin{equation*}
		\prob(\model(\individual)=1)=
		\begin{cases}
			1 &\text{if }\ \prob(\labely = 1\conditional \individual) = 1\text{ or } \prob(\labely = 1\conditional \individual)>t\\
			0 &\text{if }\ \prob(\labely = 1\conditional \individual) = 0\text{ or } \prob(\labely = 1\conditional \individual)<t\\
			q(\individual) &\text{otherwise}
		\end{cases}
	\end{equation*}
	To see that any function written in this way is Pareto-optimal, we can simply take a sequence $\vectorize{{w}}^{(k)}$ such that $\lim_{k\to\infty} {w}_1^{(k)}=t$ and apply Weller's theorem similarly as before. 
	\end{delayedproof}
\end{corollary*}
\begin{theorem*}[\ref{teo_when_works}]\Paste{teo_when_works}
	\begin{delayedproof}{teo_when_works}
	Consider a model $\model$ such that $\model\restricted{A}$ cherry-picks for a given sensitive group $A$, WLOG we can assume $A=\agroup{0}$ and 
	$
	\FEval{\model\restricted{\agroup{1}}}{}
	\leq
	\FEval{\model\restricted{\agroup{0}}}{}
	$.
	Consider now the function
	\begin{equation*}
		f(\varrate{0},\varrate{1}):=
		\FEval
		{
			\varrate{0},
			\varrate{1}
		}{\ips(\agroup{0})}
	\end{equation*}
	by fixing $E=\ips(\agroup{0})$.	By hypothesis, we have
	\begin{equation*}
		\min({\partial_{\varrate{0}} f},
		\partial_{\varrate{1}}f) \leq 0
	\end{equation*}
	As a consequence, by Theorem~\ref{teo_analisi}, the function $f$ does not strictly increase at any point $({\varrate{0}},{\varrate{1}})$. Since $\model\restricted{\agroup{0}}$ cherry-picks, we can apply Theorem~\ref{teo_atomic_cherry_picking}, and find $\varepsilon>0$ such that 
	\begin{equation*}
		\left\{\big(\jointrate{0}{\model\restricted{\agroup{0}}}{A=\agroup{0}},\jointrate{1}{\model\restricted{\agroup{0}}}{A=\agroup{0}} \big)\right\} +(0,\varepsilon)^2\cap\ips(\agroup{0})\neq \emptyset
	\end{equation*}
	In particular, we know that there exists a point $(\varfix{0},\varfix{1})$ in the above set such that 
	\begin{equation*}
	\begin{cases*}
		f(\varfix{0},\varfix{1})\geq f(\jointrate{0}{\model\restricted{\agroup{0}}}{A=\agroup{0}},\jointrate{1}{\model\restricted{\agroup{0}}}{A=\agroup{0}})
		\\
		f(\varfix{0},\varfix{1})
		\leq f(\jointrate{0}{\model\restricted{\agroup{1}}}{A=\agroup{1}},\jointrate{1}{\model\restricted{\agroup{1}}}{A=\agroup{1}})
	\end{cases*}
	\end{equation*} 
	But since $(\varfix{0},\varfix{1})$ is in the $\ips$, then we can find a model $\model'_{\agroup{0}}$ on ${\agroup{0}}$ such that 
	\begin{equation*}
	\varfix{0}=\jointrate{0}{\model'_{\agroup{0}}}{A=\agroup{0}}\text{ and }\varfix{1}=\jointrate{1}{\model'_{\agroup{0}}}{A=\agroup{0}}
	\end{equation*}
	We can now extend the model $\model'_{\agroup{0}}$ to a model $\model'$ on the full space $\featureset $ in the following way:
	\begin{equation*}
		\model'(x) =
		\begin{cases}
			\model'_{\agroup{0}}(\individual) &\text{if }\individual\in \agroup{0}\\
				\model(\individual)			&\text{otherwise}
		\end{cases}
	\end{equation*}
	We then get $\model\paretostrictless \model'$, and more importantly 
	\begin{equation*}
		\left\vert\FEval{\model'\restricted{\agroup{0}}}{}-\FEval{\model'\restricted{\agroup{1}}}{}\right\vert
		\leq 
		\left\vert\FEval{\model\restricted{\agroup{0}}}{}-\FEval{\model\restricted{\agroup{1}}}{}\right\vert
	\end{equation*}
	In particular, if we consider the set $Q$ of models that maximize the fairness problem, that is
	\begin{equation*}
		Q:=\FairnessProblem{}
	\end{equation*} 
	We can show that if $\model$ was in $Q$, then we also get $\model'\in Q$ since 
	\begin{equation*}
		\EEval{\model'}{}-\left\vert\FEval{\model'\restricted{\agroup{0}}}{}-\FEval{\model'\restricted{\agroup{1}}}{}\right\vert
		\geq 
		\EEval{\model}{}-\left\vert\FEval{\model\restricted{\agroup{0}}}{}-\FEval{\model\restricted{\agroup{1}}}{}\right\vert
	\end{equation*}
	Since we assume that $\model$ was already a maximum of the problem, it means that $\model'$ is still a maximum and the above inequality is an equality. 
	This doesn't prove yet that we can find an element in $Q$ that doesn't cherry-pick, but it proves that if $Q$ has a maximal element $\optimalmodel$ according to the Pareto order $\paretoless$, it cannot cherry-pick, otherwise we can find a strict upper bound in $Q$.
	Notice that a maximal element of $Q$ is not necessarly Pareto-optimal on $\featureset$, since maximal elements of a subset of a partially ordered set are not necessarily maximal in the whole set.\\
	The question then becomes: "Does $Q$ have maximal elements?" and we can positively answer it by applying Zorn's lemma.
	This means that if we take a sequence $\{\model^{(k)} \}_{k\in \NN}$ of models in $Q$ such that 
	\begin{equation*}
		\model^{(0)}\paretoless\model^{(1)}\paretoless\dots\paretoless\model^{(k)}\paretoless\dots
	\end{equation*} 
	we need to be able to find an upper bound in $Q$. 
	Consider a new model $\model^{(\infty)}$ 
	such that
	\begin{align*}
		\jointrate{0}{\model^{(\infty)}}{} &= \sup_{k\in \NN} \jointrate{0}{\model^{(k)}}{}\\
		\jointrate{1}{\model^{(\infty)}}{} &= \sup_{k\in \NN} \jointrate{1}{\model^{(k)}}{}
	\end{align*}
	Such model must exist since the sequences
	\begin{align*}
		\jointrate{0}{\model^{(0)}}{}\leq \jointrate{0}{\model^{(1)}}{}\leq\dots\leq \jointrate{0}{\model^{(k)}}{}\leq\dots\\
		\jointrate{1}{\model^{(0)}}{}\leq \jointrate{1}{\model^{(1)}}{}\leq\dots\leq \jointrate{1}{\model^{(k)}}{}\leq\dots
	\end{align*}
	are monotonic and the $\ips$ is compact. Also $\model^{(\infty)}$ must still be a maximum for the fairness problem due to the continuity of $\EEval{}{}$ and $\FEval{}{}$. This means that $\model^{(\infty)}$ is an upper bound for the sequence. The only other property we need to prove is that $Q$ is not empty, which follows from the fact that the $\ips$ is a compact set and the fairness problem is a continuous one. So $Q$ has maximal elements, which again means that there exists $\optimalmodel $ that satisfies the claim.
\end{delayedproof}
\end{theorem*}
\end{document}